\documentclass[runningheads]{llncs}
\usepackage[T1]{fontenc}
\usepackage{lmodern}

\usepackage{graphicx}
\usepackage[title]{appendix}

\usepackage{hyperref}
\hypersetup{
  colorlinks = true,
  citecolor = blue,
  linkcolor = blue,
  urlcolor = blue
}

\usepackage{cite}
\usepackage{multirow} 
\usepackage[ruled,vlined,linesnumbered,noend]{algorithm2e}
\usepackage{color}
\usepackage{xcolor}
\usepackage{multirow}
\usepackage{adjustbox}
\usepackage{graphicx} 
\usepackage{amsmath}   
\usepackage{amssymb}
\usepackage{mathtools}
\usepackage{microtype}
\usepackage{booktabs}
\usepackage{orcidlink}

\usepackage{tikz}
\usepackage{pgfplots}
\usepackage[linesnumbered,ruled,vlined]{algorithm2e}

\usepackage{xcolor}

 \title{A Linear Weight Transfer Rule for Local Search}
 \author{Md Solimul Chowdhury \orcidlink{0000-0001-8429-2108} \and
 Cayden R. Codel \orcidlink{0000-0003-3588-4873} \and
 Marijn J.H. Heule \orcidlink{0000-0002-5587-8801}}

\authorrunning{Chowdhury et al.}
\institute{Carnegie Mellon University, Pittsburgh, Pennsylvania, United States \\
\email{\{mdsolimc, ccodel, mheule\}@cs.cmu.edu}}
\date{}

\newcommand{\F}{F}
\newcommand{\winit}{w_0}

\newcommand{\linrevs}{\texttt{lw-ith}}
\newcommand{\linorigs}{\texttt{lw-itl}}
\newcommand{\lineqs}{\texttt{lw-ite}}

\newcommand{\wrv}{\texttt{wrv}}
\newcommand{\sv}{\texttt{sv}}

\newcommand{\mphf}{$\mathtt{mphf}$}
\newcommand{\mm}{$\mathtt{MM}$}
\newcommand{\wap}{$\mathtt{wap}$}
\newcommand{\asias}{$\mathtt{asias}$}
\newcommand{\ptn}{$\mathtt{ptn}$}
\newcommand{\twentysix}{$\mathtt{26x26}$}

\newcommand{\greens}{$\mathtt{vdw}$}

\newcommand{\weight}{W}

\newcommand{\paws}{\textsc{paws}}
\newcommand{\saps}{\textsc{saps}}
\newcommand{\ddfw}{\textsc{ddfw}}
\newcommand{\probsat}{\textsc{probsat}}
\newcommand{\ubcsat}{\textsc{ubcsat}}
\newcommand{\yalsat}{\texttt{yalsat}}
\newcommand{\yalsatddfw}{\texttt{yal-lin}}
\newcommand{\yalsatprob}{\texttt{yal-prob}}

\newcommand{\parama}{\texttt{a}}
\newcommand{\paramc}{\texttt{c}}
\newcommand{\agt}{\ifmmode \parama_{>} \else \parama$_{>}$ \hspace{-4pt} \fi}
\newcommand{\ale}{\ifmmode \parama_{=} \else \parama$_{=}$ \hspace{-4pt} \fi}
\newcommand{\cgt}{\ifmmode \paramc_{>} \else \paramc$_{>}$ \hspace{-4pt} \fi}
\newcommand{\cle}{\ifmmode \paramc_{=} \else \paramc$_{=}$ \hspace{-4pt} \fi}

\newcommand{\uvars}{\texttt{uvars}}

\newcommand{\pmbest}{\texttt{grdy}}

\newcommand{\pmwrand}{\texttt{wrnd}}
\newcommand{\spt}{\texttt{spt}}

\newcommand{\maxtries}{MAXTRIES} 
\newcommand{\maxflips}{MAXFLIPS} 
\newcommand{\neighbors}{\mathrm{Neighbors}}

\newcommand{\intwt}{\texttt{fixedwt}}
\newcommand{\intwtshort}{\texttt{fw}}
\newcommand{\linwt}{\texttt{linearwt}}
\newcommand{\linwtshort}{\texttt{lw}}

\newcommand{\cspt}{\texttt{cspt}}

\newcommand{\ubcsatddfw}{\texttt{ubc-ddfw}}
\newcommand{\instancesA}{COMB}
\newcommand{\instancesB}{SATComp}

\definecolor{cola}{rgb}{0.878431, 0.235294, 0.192157}
\definecolor{colb}{rgb}{0.552941, 0.72549, 0.792157}
\definecolor{colc}{rgb}{0.964706, 0.745098, 0}
\definecolor{cold}{rgb}{0.917647, 0.462745, 0}
\definecolor{cole}{rgb}{0.54902, 0.509804, 0.47451}
\definecolor{colf}{rgb}{0.643137, 0.858824, 0.909804}
\definecolor{colg}{rgb}{0.141176, 0.313725, 0.603922}
\definecolor{colh}{rgb}{0.709804, 0.741176, 0}
\definecolor{coli}{rgb}{0.835294, 0, 0.196078}
\definecolor{colj}{rgb}{0, 0.592157, 0.662745}
\definecolor{colk}{rgb}{0.67451, 0.0784314, 0.352941}
\definecolor{coll}{rgb}{0.333333, 0.313725, 0.145098}
\definecolor{colm}{rgb}{0.396078, 0.113725, 0.196078}
\definecolor{coln}{rgb}{0.294118, 0.219608, 0.298039}
\definecolor{colo}{rgb}{0, 0.239216, 0.298039}
\definecolor{colp}{rgb}{0.305882, 0.211765, 0.160784}
\definecolor{colq}{rgb}{0.560784, 0.6, 0.243137}
\definecolor{colr}{rgb}{0.576471, 0.152941, 0.172549}
\definecolor{colt}{rgb}{0.313725, 0.027451, 0.470588}
\definecolor{colu}{rgb}{0, 0.156863, 0.333333}
\definecolor{colv}{rgb}{0.776471, 0.690196, 0.737255}
\definecolor{colw}{rgb}{0.733333, 0.772549, 0.572549}
\definecolor{colx}{rgb}{0.839216, 0.823529, 0.768627}

\newcommand{\marka}{triangle}
\newcommand{\markb}{triangle*}
\newcommand{\markc}{square}
\newcommand{\markd}{square*}
\newcommand{\marke}{+}
\newcommand{\markf}{o}
\newcommand{\markg}{*}
\newcommand{\markh}{x}
\newcommand{\marki}{diamond}
\newcommand{\markj}{diamond*}
\newcommand{\markk}{pentagon}

\newcommand{\solvera}{\large \tt fw-c.01-grdy}
\newcommand{\solverb}{\large \tt fw-c.01-wrnd}
\newcommand{\solverc}{\large \tt fw-c.1-grdy}
\newcommand{\solverd}{\large \tt fw-c.1-wrnd}
\newcommand{\solvere}{\large \tt \lineqs-c.1-grdy}
\newcommand{\solverf}{\large \tt \lineqs-c.1-wrnd}
\newcommand{\solverg}{\large \tt \linrevs-c.1-grdy}
\newcommand{\solverh}{\large \tt \linrevs-c.1-wrnd}
\newcommand{\solveri}{\large \tt \linorigs-c.1-grdy}
\newcommand{\solverj}{\large \tt \linorigs-c.1-wrnd}
\newcommand{\solverk}{\large \tt yal-prob}

\begin{document}

\maketitle

\begin{abstract}
  The \emph{Divide and Distribute Fixed Weights} algorithm (\ddfw{}) is a dynamic local search SAT-solving algorithm that transfers weight from satisfied to falsified clauses in local minima.
  \ddfw{} is remarkably effective on several hard combinatorial instances. Yet, despite its success, it has received little study since its debut in 2005.
  In this paper, we propose three modifications to the base algorithm:
  a linear weight transfer method that moves a dynamic amount of weight between clauses in local minima,
  an adjustment to how satisfied clauses are chosen in local minima to give weight,
  and a weighted-random method of selecting variables to flip.
  We implemented our modifications to \ddfw{} on top of the solver \yalsat{}.
  Our experiments show that our modifications boost the performance compared to the original \ddfw{} algorithm on multiple benchmarks, including those from the past three years of SAT competitions.
  Moreover, our improved solver exclusively solves hard combinatorial instances that refute a conjecture on the lower bound of two Van der Waerden numbers set forth by Ahmed et al. (2014), and it performs well on a hard graph-coloring instance that has been open for over three decades.
\end{abstract}

\section{Introduction}

Satisfiability (SAT) solvers are powerful tools, able to efficiently solve problems from a broad range of applications such as verification~\cite{HardwareDesign}, encryption~\cite{cms}, and planning~\cite{SATplan,allUIP}.
The most successful solving paradigm is conflict-driven clause learning (CDCL)~\cite{grasp,Chaff}.
However, stochastic local search (SLS) outperforms CDCL on many classes of satisfiable formulas~\cite{gsat,walksat,ubcsatthesis,LiLi,CCAnr}, and it can be used to guide CDCL search~\cite{relaxed22}.  

SLS algorithms solve SAT instances by incrementally changing a truth assignment until a solution is found or until timeout.
At each step, the algorithm flips the truth value of a single boolean variable, often according to some heuristic.
A common heuristic is flipping variables that reduce the number of falsified clauses in the formula, but this is not the only one.
The algorithm reaches a \emph{local minimum} when no variable can be flipped to improve its heuristic.
At that point, the algorithm either adjusts its truth assignment or internal state to \emph{escape} the local minimum, or it starts over.
Refer to chapter 6 from the Handbook of Satisfiability~\cite{HandbookSAT} for a more detailed discussion of SLS algorithms.

\emph{Dynamic local search} (DLS) algorithms are SLS algorithms that assign a weight to each clause.
They then flip variables to reduce the amount of weight held by the falsified clauses.
DLS algorithms escape local minima by adjusting clause weights until they can once again flip variables to reduce the amount of falsified weight.

Several DLS algorithms have been studied.
For example, the Pure Additive Weighting Scheme algorithm (\paws)~\cite{paws} and the Scaling and Probabilistic Smoothing algorithm (\saps)~\cite{saps} both increase the weight of falsified clauses in local minima.
A drawback of this method of escaping local minima is that the clause weights must periodically be re-scaled to prevent overflow.

The Divide and Distribute Fixed Weights algorithm (\ddfw{})~\cite{ddfw} introduces an alternative way of escaping local minima: increase the weight of falsified clauses by taking weight from satisfied clauses.
In local minima, \ddfw{} moves a fixed, constant amount of weight to each falsified clause from a satisfied clause it shares at least one literal with.
The transfer method keeps the total amount of clause weight constant, eliminating the need for a re-scaling phase.
Another consequence of this transfer method is that as more local minima are encountered, difficult-to-satisfy clauses gather more weight.
Thus, \ddfw{} dynamically identifies and prioritizes satisfying hard clauses.

Recent work shows that \ddfw{} is an effective algorithm.
For example, \ddfw{} (as implemented in {\ubcsat}~\cite{ubcsat}\footnote{To the best of our knowledge, there is no official implementation or binary of original \ddfw{} \cite{ddfw} available.}) is remarkably effective on matrix multiplication and graph-coloring problems~\cite{matrix,c2c}.
Yet despite its success, \ddfw{} has received little research attention.
In this paper, we revisit the \ddfw{} algorithm to study why it works well and to improve its performance.

Our contributions are as follows.
We propose three modifications to the \ddfw{} algorithm.
We first introduce a linear weight transfer rule to allow for a more dynamic transfer of weight in local minima.
We then adjust a performance-critical parameter that randomizes which satisfied clause gives up weight in local minima.
Our adjustment is supported by an empirical analysis.
Finally, we propose a new randomized method for selecting which variable to flip.
We implement each of our modifications on top of the state-of-the-art SLS solver \yalsat{} to create a new implementation of \ddfw{} that supports parallelization and restarts.
We then evaluate our solver against a set of challenging benchmarks collected from combinatorial problem instances and the past three years of SAT competitions.
Our results show that our modifications boost the performance of \ddfw{}:
Our best-performing version of \ddfw{} solves 118 SAT Competition instances, a vast improvement over a baseline of 83 solves from the original algorithm.
Our solver also exhibits a 16\% improvement over the baseline on a set of combinatorial instances.
Moreover, in parallel mode, our solver solves instances that refute a conjecture on the lower bound of two van der Waerden numbers~\cite{kullman}, and it matches performance with the winning SLS solver from the 2021 SAT competition on a graph-coloring instance that has been open for the past three decades.

\section{Preliminaries}\label{sec:preliminaries}

SAT solvers operate on propositional logic formulas in \emph{conjunctive normal form} (CNF).
A CNF formula~$\F{} = \bigwedge_i C_i$ is a conjunction of clauses, and each clause~$C_i = \bigvee_j \ell_j$ is a disjunction of boolean literals.
We write $v$ and $\overline{v}$ as the positive and negative literals for the boolean variable $v$, respectively.

A truth assignment~$\alpha$ maps boolean variables to either true or false.
A literal~$v$ (resp. $\overline v$) is satisfied by $\alpha$ if $\alpha(v)$ is true ($\alpha(v)$ is false, respectively).
A clause $C$ is satisfied by $\alpha$ if $\alpha$ satisfies at least one of its literals.
A formula $\F{}$ is satisfied by $\alpha$ exactly when all of its clauses are satisfied by $\alpha$.
Two clauses $C$ and $D$ are \emph{neighbors} if there is a literal $\ell$ with $\ell \in C$ and $\ell \in D$.
Let $\neighbors(C)$ be the set of neighbors of $C$ in $\F{}$, excluding itself.

Many SLS algorithms assign a weight to each clause.
Let $\weight : \mathcal{C} \rightarrow \mathbb{R}_{\geq 0}$ be the mapping that assigns weights to the clauses in $\mathcal{C}$.
One can think of $\weight(C)$ as the cost to leave $C$ falsified.
We call the total amount of weight held by the falsified clauses, the \emph{falsified weight}.
A variable that, when flipped, reduces the falsified weight is called a \emph{weight-reducing variable} (\wrv).
A variable that doesn't affect the falsified weight when flipped is a \emph{sideways variable} (\sv).

\section{The \ddfw{} Algorithm}\label{sec:ddfw}

Algorithm~\ref{alg:ddfw} shows the pseudocode for the \ddfw{} algorithm.
\ddfw{} attempts to find a satisfying assignment for a given CNF formula $\F$ over \maxtries{} trials.
The weight of each clause is set to $\winit$ at the start of the algorithm.
Each trial starts with a random assignment.
By following a greedy heuristic method, \ddfw{} selects and then flips weight-reducing variables until none are left.
At this point, it either flips a sideways variable, if one exists and if a weighted coin flip succeeds, or it enters the weight transfer phase, where each falsified clause receives a fixed amount of weight from a maximum-weight satisfied neighbor.
Occasionally, \ddfw{} transfers weight from a random satisfied clause instead, allowing weight to move more fluidly between neighborhoods.
The amount of weight transferred depends on whether the selected clause has more than $\winit$ weight.

\begin{algorithm} 
\DontPrintSemicolon
\SetAlgoNoLine
\SetInd{0.48em}{0.11em}

\KwIn{CNF Formula $\F$, $\winit$, $\spt$, $\cspt$, \cgt, \cle}
\KwOut{Satisfiability of $\F$}
\SetKwBlock{Begin}{function}{end function}
{
  $\weight(C) \leftarrow \winit$ for all $C \in \F$\;
  \For{$t=1$ to \maxtries}
  {
    $\alpha \leftarrow $ random truth assignment on the variables in $\F$\;
    \For{$f=1$ to \maxflips}
    {
      \lIf{$\alpha$ satisfies $\F$}
        {return ``SAT''}
      \Else
      { 
        \uIf{there is a \wrv} 
        {
          Flip a \wrv{} that most reduces the falsified weight\;
        }
        \uElseIf {there is a \sv{} and rand $\leq \spt$}
        {
          Flip a sideways variable\;
        }
        \uElse {
            \ForEach{falsified clause $C$}
            {
                $C_s \leftarrow$ maximum-weighted satisfied clause in $\neighbors(C)$\;\color{black}
                \uIf{$\weight (C_s) < \winit$ \textnormal{ or} rand $\leq \cspt$}
                {
                    $C_s \leftarrow$ random satisfied clause with $\weight \geq \winit$\;  
                }
                \uIf {$\weight (C_s) > \winit$}
                {
                    Transfer \cgt weight from $C_s$ to $C$\;
                }
                \Else
                {
                    Transfer \cle weight from $C_s$ to $C$\;
                }
            }
        } 
        }
    }
  }\label{endfor}
  return ``No SAT''\;
}
\caption{The \ddfw{} algorithm}\label{alg:ddfw}
\end{algorithm}

There are five parameters in the original \ddfw{} algorithm: the initial weight~$\winit$ given to each clause, the two weighted-coin thresholds \spt{} and \cspt{} for sideways flips and transfers from random satisfied clauses, and the amount of weight to transfer in local minima \cgt and \cle.
In the original \ddfw{} paper, these five values are fixed constants, with $\winit = 8$, $\spt{} = 0.15$, $\cspt{} = 0.01$, $\cgt = 2$, and $\cle = 1$.

\ddfw{} is unique in how it transfers weight in local minima.
Similar SLS algorithms increase the weight of falsified clauses (or decrease the weight of satisfied clauses) globally;
weight is added and removed based solely on whether the clause is satisfied.
\ddfw{} instead moves weight among clause neighborhoods, with falsified clauses receiving weight from satisfied neighbors.

One reason why this weight transfer method may be effective is that satisfying a falsified clause $C$ by flipping literal $\overline{\ell}$ to $\ell~(\in C)$ increases the number of true literals in satisfied clauses that neighbor $C$ on $\ell$.
Thus, $C$ borrows weight from satisfied clauses that tend to remain satisfied when $C$ itself becomes satisfied.
As a result, \ddfw{} satisfies falsified clauses while keeping satisfied neighbors satisfied.

The existence of \emph{two} weight transfer parameters \cgt and \cle deserves discussion.
Let \emph{heavy clauses} be those clauses $C$ with $\weight(C) > \winit$.
Lines 16-19 in Algorithm~\ref{alg:ddfw} allow for a different amount of weight to be taken from heavy clauses than from clauses with the initial weight.
Because lines 14-15 ensure that the selected clause $C_s$ will have at least $\winit$ weight, \cle is used when $W(C_s) = \winit$ and \cgt is used when $W(C_s) > \winit$ (hence the notation).
The original algorithm sets $\cgt = 2$ and $\cle = 1$, which has the effect of taking more weight from heavy clauses.


\section{Solvers, Benchmarks, and Hardware}

The authors of the original \ddfw{} algorithm never released their source code or any binaries.
The closest thing we have to a reference implementation is the one in the SLS SAT-solving framework \ubcsat{}~\cite{ubcsatthesis,ubcsat}.
We call this implementation \ubcsatddfw, and we use it as a baseline in our experiments.

Unfortunately, \ubcsatddfw{} cannot be extended to implement our proposed modifications due to its particular architecture.
Instead, we implemented \ddfw{} on top of \yalsat{}~\cite{yalsat}, which
is currently one of the strongest local search SAT solvers. For example, it is the only local search solver in Mallob-mono~\cite{mallob}, the clear winner of the cloud track in the SAT Competitions of 2020, 2021, and 2022.
\yalsat{} uses \probsat{}~\cite{probsatphd} as its underlying algorithm, which flips variables in falsified clauses drawn from an exponential probability distribution.

One benefit of implementing \ddfw{} on top of \yalsat{} is that is \yalsat{} supports parallelization, which can be helpful when solving challenging formulas.
In our experiments, we compare our implementation of \ddfw{} to \ubcsatddfw{} to verify that the two implementations behave similarly.

Our implementation of \ddfw{} on top of \yalsat{} was not straightforward.
First, we switched the underlying SLS algorithm from \probsat{} to \ddfw.
Then we added additional data structures and optimizations to make our implementation efficient.
For example, one potential performance bottleneck for \ddfw{} is calculating the set of weight-reducing variables for each flip.
Every flip and adjustment of clause weight can change the set, so the set must be re-computed often.
A naive implementation that loops through all literals in all falsified clauses is too slow, since any literal may appear in several falsified clauses, leading to redundant computation.
Instead, we maintain a list of variables \uvars{} that appear in any falsified clause.
After each flip, this list is updated.
To compute the set of weight-reducing variables, we iterate over the variables in \uvars, hitting each literal once.
In this way, we reduce redundant computation.

Adding our proposed modifications to our implementation was simpler.
We represent clause weights with floating-point numbers, and the linear weight transfer rule replaced the original one.
We also made the variable selection and weight transfer methods modular, so our modifications slot in easily\footnote{Source code of our system is available at \url{https://github.com/solimul/yal-lin} \color{black}}.


We evaluated our implementations of \ddfw{} against two benchmarks.
The \textbf{Combinatorial (\instancesA{})} set consists of 65 hard instances from the following eight benchmarks families collected by Heule:\footnote{\url{https://github.com/marijnheule/benchmarks}}
(i) \twentysix{} (4 grid positioning instances), (ii) \asias{} (2 almost square packing problems), (iii) \mm{} (20 matrix multiplication instances), (iv) \mphf{} (12 cryptographic hash instances), (v) \ptn{} (2 Pythagorean triple instances), (vi) Steiner (3 Steiner triples cover instances~\cite{Steiner}),  (vii) \wap{} (9 graph-coloring instances~\cite{colorings}),  and (viii) \greens{} (13 van der Waerden number instances).
These benchmarks are challenging for modern SAT solvers, including SLS solvers.
The \wap{} benchmark contains three instances that have been open for three decades, and \greens{} contains two instances that, if solved, refute conjectures on lower-bounds for two van der Waerden numbers~\cite{kullman}.

The \textbf{SAT Competition (\instancesB{})} set consists of all 1,174 non-duplicate main-track benchmark instances from the 2019 SAT Race and the 2020 and 2021 SAT Competitions. The competition suites contain medium-hard to very challenging benchmarks, most of which are contributed by the competitors.

Unless otherwise specified, we used a timeout of 18,000 and 5,000 seconds for the \instancesA{} and \instancesB{} instances, respectively, in our experiments.

We used the StarExec cluster~\cite{starexec}, where each node has an Intel CPU E5 CPU with a 2.40 GHz clock speed and a 10240 KB cache.
For experiments in this cluster, we used at most 64 GB of RAM.
To perform experiments on the 3 open \wap{} and 2 \greens{} instances, we used a different cluster with the following specifications:
two AMD EPYC 7742 CPUs with a 3.40 GHz clock speed, each with 64 cores, 256 MB of L3 cache, and 512 GB of RAM.

\section{Modifications to the \ddfw{} Algorithm}

We propose three modifications to \ddfw.
The first is a linear rule for transferring a dynamic amount of weight in local minima.
The second is an adjustment of the \cspt{} parameter.
The third is the introduction of a weighted-random method for selecting which variable to flip.

\subsection{The Linear Weight Transfer Rule}

The reference implementation of \ddfw, \ubcsatddfw, represents its clause weights as integers and transfers fixed integer weights in local minima.
While this design decision allows \ubcsatddfw{} to have a fast implementation, it unnecessarily restricts the amount of weight transferred in local minima to be integer-valued.
In addition, the choice to transfer a fixed, constant amount of weight prevents \ddfw{} from adapting to situations where more weight must be transferred to escape a local minimum, thus requiring multiple weight transfer rounds.
To address these concerns, we propose a dynamic linear weight transfer rule to operate on floating-point-valued clause weights.

Let $C_S$ be the selected satisfied clause from which to take weight in a local minimum, as in line 13 in Algorithm~\ref{alg:ddfw}.
Our new rule transfers $$\parama{} * \weight (C_S) + \paramc{}$$ weight, where $0 \leq \parama{} \leq 1$ is a multiplicative parameter and $\paramc{} \geq 0$ is an additive parameter.

\begin{figure}[h]
  
    \includegraphics[width=\linewidth]{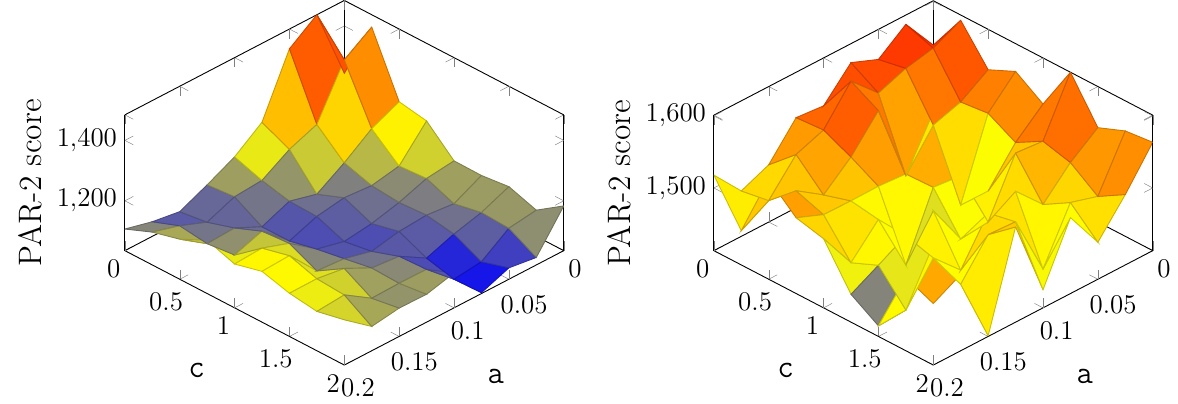}
  \caption{Parameter searches for $\parama \in [0, 0.2]$ in steps of 0.05 and $\paramc \in [0, 2]$ in steps of 0.25 on the \instancesB{} (left plot) and \instancesA{} (right plot) instances. A lower PAR-2 score is better. There is not a datum for $(\parama, \paramc) = (0, 0)$ since no weight would be transferred.}
  \label{fig:ac_landscape}
\end{figure}

It is not clear that the addition of a multiplicative parameter is helpful, nor what a good pair of $(\parama, \paramc)$ values would be.
So, we performed a parameter search with our solver for $\parama \in [0, 0.2]$ in steps of 0.05 and $\paramc \in [0, 2]$ in steps of 0.25 for both of our instance sets with a 900 second timeout per run.
(A parameter search using all 1,174 instances in the \instancesB{} set was not feasible. We instead did the search on the 168 instances from \instancesB{} set that were solved by some setting in earlier experimentation.
In Section~\ref{sec:eval}, all instances are used.)
The PAR-2 scores\footnote{The PAR-2 score is defined as the average solving time, while taking 2 * \texttt{timeout} as the time for unsolved instances. A lower score is better.} for the \instancesB{} and \instancesA{} benchmark sets for each pair of $(\parama, \paramc)$ values are shown in Figure~\ref{fig:ac_landscape}.

The plots in Figure~\ref{fig:ac_landscape} show that values of \parama{} and \paramc{} close to 0 degrade performance, likely due to the need for many weight-transfer rounds to escape local minima.
The beneficial effect of higher values of \parama{} and \paramc{} is more pronounced in the parameter search on the \instancesB{} instances (the left plot).
Since the best-performing settings have nonzero \parama{} and \paramc{} values, we infer that both parameters are needed for improved performance.

\subsection{How Much Weight Should be Given Away Initially?}\label{subsec:linwt}

On lines 16-19 of Algorithm~\ref{alg:ddfw}, \ddfw{} takes \cgt weight away from the selected clause~$C_s$ if $C_s$ is heavy and \cle weight otherwise.
The linear rule introduced above can similarly be extended to four parameters: $\parama_{>}$, $\parama_{=}$, $\paramc_{>}$, and $\paramc_{=}$. 

In the original \ddfw{} paper, \cgt (= 2) is greater than \cle (= 1), meaning that heavy clauses give away more weight than clauses with the initial weight in local minima.
The intuition behind this is simple: clauses with more weight should give away more weight.
For the extended linear rule, one could adopt a similar strategy by setting \agt greater than \ale and \cgt greater than $\paramc_{=}$. 

However, one effect of our proposed linear rule is that once clauses give or receive weight, they almost never again have exactly $\winit$ weight.
As a result, the parameters \ale{} and \cle{} control how much weight a clause gives away \emph{initially}.
Since the maximum-weight neighbors of falsified clauses tend to be heavy as the search proceeds, the effect of \ale and $\paramc_{=}$ diminishes over time, but they remain important at the start of the search and for determining how much weight the algorithm has available to assign to harder-to-satisfy clauses.
The findings in a workshop paper~\cite{ddfwpos} by two co-authors of this paper indicate that \ddfw{} achieves a better performance when clauses initially give more weight.
These findings suggest setting \cle greater than \cgt and \ale greater than $\parama_{>}$.
In Section~\ref{sec:eval}, we evaluate \ddfw{} on the extended linear rule and investigate whether clauses should initially give away more or less weight.

\subsection{The \cspt{} Parameter}\label{subsec:cspt}

On lines 14-15 of Algorithm~\ref{alg:ddfw}, \ddfw{} sometimes discards the maximum-weight satisfied neighboring clause~$C_s$ and instead selects a random satisfied clause.
The \cspt{} parameter controls how often the weighted coin flip on line 14 succeeds.
Though these two lines may appear to be minor, a small-scale experiment revealed that the \cspt{} parameter is performance-critical.
We ran our implementation of the original \ddfw{} algorithm on the COMB set with an 18,000 second timeout.
When we set \cspt{} to 0, meaning that falsified clauses received weight solely from satisfied neighbors, it solved a single instance;
when we set \cspt{} to 0.01 (the value in the original \ddfw{} algorithm), it solved 21 instances.

\begin{figure}[t]
  \centering
  \begin{tikzpicture}
    \begin{axis}[
      title = {},
      xlabel = {\cspt{} value},
      ylabel = {PAR-2 score},
      xmin = 0,
      xmax = 1,
      width=\textwidth,
      height=120pt,
      grid = major,
      legend entries = {},]
      \addplot[only marks,mark size=.8pt, color=blue] table {data/wap-average-par2.tsv};
    \end{axis}
  \end{tikzpicture}
  \caption{The impact of \cspt{} values on the performance of \ddfw{} on the \wap{} instances.}
  \label{fig:ds-cspt}
\end{figure}
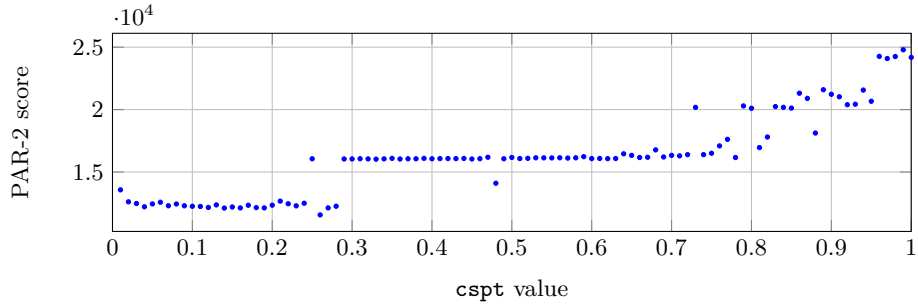

Among the eight families in \instancesA{}, the \wap{} family was the most sensitive to the change of \cspt{} value from 0 (solved 0) to 0.01 (solved 6 out of 9).
We isolated these nine instances and ran a parameter search on them for $\cspt{} \in [0.01, 1]$ in steps of 0.01, for a total of 900 runs.
We used an 18,000 second timeout per run.
The PAR-2 scores are reported in Figure~\ref{fig:ds-cspt}.

In Figure~\ref{fig:ds-cspt}, we observe that \cspt{} values near 0 and above 0.2 cause an increase in the PAR-2 score.
These results indicate that \ddfw{} is sensitive to the \cspt{} value and that the \cspt{} value should be set higher than its original value of 0.01, but not too high, which could potentially degrade the performance of the solver. We use these observations to readjust the \cspt{} parameter in our empirical evaluation presented in Section~\ref{sec:eval}.

\subsection{A Weighted-random Variable Selection Method}

On line 8 of Algorithm~\ref{alg:ddfw}, \ddfw{} flips a weight-reducing variable that most reduces the amount of falsified weight.
Such a greedy approach may prevent \ddfw{} from exploring other, potentially better areas of the search space.
Inspired by \probsat{}, which makes greedy moves only some of the time, we introduce a new randomized method that flips a weight-reducing variable according to the following probability distribution:
$$\mathbb{P} (\text{Flipping \wrv{} } v) = \frac{\Delta \weight(v)}{\sum_{v \in \wrv} \Delta \weight (v)},$$ where $\Delta \weight (v)$ is the reduction in falsified weight if $v$ is flipped.


\section{Empirical Evaluation}\label{sec:eval}

In this section, we present our empirical findings.
Since we evaluated several different solvers, we refer to the solvers by the following names:
the \ubcsat{} version of \ddfw{} is \ubcsatddfw{},
the version of \yalsat{} that implements \probsat{} is \yalsatprob{},
and our implementation of \ddfw{} on top of \yalsat{} is \yalsatddfw{}.
In all of our experiments, we use the default random seed\footnote{Results for additional experiments with a different seed is available in Appendix \ref{appendix:A}} present in each solver, and we set the initial clause weight $\winit{} = 8$, as in the original \ddfw{} paper.


In our experiments with \yalsatddfw{}, we varied the configuration of the solver according to our proposed modifications.
We use the identifying string \texttt{W-cC-P} to refer to a configuration for \yalsatddfw{}, where $\texttt{W} \in \{ \intwtshort{}, \linwtshort{} \}$ is the weight transfer method (\intwtshort{} stands for ``fixed weight,'' \linwtshort{} for ``linear weight''), $\texttt{C} \in \{ 0.01, 0.1 \}$ is the \cspt{} value, and $\texttt{P} \in \{ \pmbest{}, \pmwrand{} \}$ is the variable selection method (\pmbest{} stands for the original ``greedy'' method, and \pmwrand{} stands for our proposed ``weighted random'' method).
For example, the string \texttt{\intwtshort-c.01-\pmbest} describes the original \ddfw{} algorithm, with $\cgt = 2$ and $\cle = 1$.

\subsection{Evaluation Without Restarts}\label{sec-withoutrestart}

We evaluate how \yalsatddfw{} performs without restarts, meaning that \ddfw{} runs until timeout without starting from a fresh random assignment.
To disable restarts, we set \maxtries{} to 1 and \maxflips{} to an unlimited number of flips.
For the \instancesA{} and \instancesB{} benchmark sets, we set a timeout of 18,000 and 5,000 seconds, respectively.


We first checked that our solver \yalsatddfw{} (with configuration \texttt{\intwtshort-c.01-\pmbest}) behaves similarly to the baseline implementation, \ubcsatddfw{}.
The solvers performed almost identically on the two benchmark sets: \ubcsatddfw{} solved 22 of the \instancesA{} instances and 80 of the \instancesB{} instances; \yalsatddfw{} solved 21 and 83, respectively.
We attribute the slight difference in solve counts to random noise.
These results indicate that we implemented \yalsatddfw{} correctly.

We next evaluate how \yalsatddfw{} performs under changes in the \cspt{} value and variable selection method.
We run \yalsatddfw{} with the fixed weight transfer method on both benchmarks with all four combinations of $\texttt{C} \in \{ 0.01, 0.1 \}$ and $\texttt{P} \in \{ \pmbest{}, \pmwrand{} \}$.
The solve counts and PAR-2 scores are shown in Table~\ref{tab:segment3}.

Isolating just the change in variable selection method (scanning across rows in Table~\ref{tab:segment3}), we see that the weighted-random method outperforms the greedy method for each benchmark and \cspt{} value.
There is improvement both in the solve count (ranging from an additional 1 to 5 solves) and in the PAR-2 score.
While the improvements may be random noise, the results indicate that injecting some randomness into how variables are flipped may lead to better performance.

Isolating the change in \cspt{} value (scanning down columns in Table~\ref{tab:segment3}), we see that the higher \cspt{} value of 0.1 outperforms the \cspt{} value of 0.01.
Improvements range from 1 additional solve to 16 additional solves.
We note that the improvements when increasing the \cspt{} value are more pronounced than when changing the variable selection method, which gives further evidence that the \cspt{} value is performance-critical.
In Section~\ref{sec:conclusion}, we present a possible explanation for why the \cspt{} parameter is so important.

\begin{table}[t]
  \caption{Solve counts and PAR-2 scores for different configurations of \yalsatddfw{}. The configurations vary the \cspt{} value and the variable selection method, with the weight transfer method being \intwtshort. The best configuration for each benchmark is bolded.}
  \centering
  \label{tab:segment3}
  \begin{tabular}{c|cc|cc|cc|cc}
    \toprule
     & \multicolumn{4}{c|}{\instancesA}                                                                                                & \multicolumn{4}{c}{\instancesB}                                                                                                            \\ 
          {\cspt{}}                           & \multicolumn{2}{c|}{\pmbest}                                              & \multicolumn{2}{c|}{\pmwrand}                        & \multicolumn{2}{c|}{\pmbest}                                               & \multicolumn{2}{c}{\pmwrand}         \\ 
           {value}                          & {~\#solved~}    & {~PAR-2~}            & {~\#solved~}    & {~PAR-2~}           & {~\#solved~}    & {~PAR-2~}             & {~\#solved~}     & {~PAR-2~} \\ \midrule
    ~~0.01~~                        & {21}          & {25393}          & {24}          & 23871          & {83}          & {9339}          & {87}           & 9312                   \\ 
    ~~0.1~~                         & {24} & {23137} & {\textbf{25}} & \textbf{22538} & {98} & {9223} & {\textbf{103}} & \textbf{9188}          \\ \bottomrule
  \end{tabular}
\end{table}

\vspace{.1cm}
\noindent \textbf{The linear weight transfer rule.}
As we noted in Section~\ref{subsec:linwt}, the linear weight transfer rule can be extended to include four parameters: two multiplicative and two additive.
We tested \yalsatddfw{} on three particular settings of these four parameters, which we call 
\linorigs{} (\textbf{l}inear \textbf{w}eight \textbf{i}nitial \textbf{t}ransfer \textbf{l}ow), 
\linrevs{} (\textbf{l}inear \textbf{w}eight \textbf{i}nitial \textbf{t}ransfer \textbf{h}igh), 
and \lineqs{} (\textbf{l}inear \textbf{w}eight \textbf{i}nitial \textbf{t}ransfer \textbf{e}qual).
\begin{itemize}
    \item \linorigs{} takes a low initial transfer from clauses in local minima by setting \ale < \agt and \cle < \cgt.
    
    \item \linrevs{} takes a high initial transfer from clauses in local minima by setting \ale > \agt and \cle > \cgt.
    
    \item \lineqs{} does not distinguish clauses by weight, and sets the two pairs of parameters equal.
\end{itemize}

In the left plot of Figure \ref{fig:ac_landscape}, \parama{} values for the top 10\% of the settings (by PAR-2 scores) are in the range [0.05, 0.1].
Hence, we use 0.05 and 0.1 as the values for \agt and \ale in \linorigs{} and \linrevs{}.
We keep the values for \cgt and \cle at 2 and 1, following the original \ddfw{} algorithm.
For \lineqs{}, we take the average of the two pairs of values, with $\parama_{>} = \parama_{=} = 0.075$ and $\paramc_{>} = \paramc_{=} = 1.75$\color{black}.
Table~\ref{tab:linparam} shows the parameter values for the three configurations that we tested.

\begin{table}[]
  \caption{Parameter values for three versions of \linwt{}}
  \label{tab:linparam}
  \centering
  \begin{tabular}{@{~~~}c@{~~~}|@{~~~}c@{~~~}|@{~~~}c@{~~~}|@{~~~}c@{~~~}|@{~~~}c@{~~~}}
    \hline
    
    \toprule
    
    \linwt{} versions & {\agt}  & \ale  & {\cgt} & \cle \\ \midrule

    \linorigs{}                                   & {0.1}  & 0.05 & {2}   & 1   \\  
    \lineqs{}                                 & {0.075} & 0.075  & {{1.75}}   &  1.75  \\ 
    \linrevs{}                                 & {0.05} & 0.1  & {1}   & 2   \\ \bottomrule
  \end{tabular}
\end{table}

We compare our three new configurations against the original one across the two variable selection methods.
We set \cspt{} = 0.1, as our prior experiment showed it to be better than 0.01.
Table~\ref{tab:segment4} summarizes the results.

\begin{table}[]
  \centering
  \caption{Solve counts and PAR-2 scores for different configurations of \yalsatddfw{}. The configurations vary the linear weight transfer method while keeping the \cspt{} value fixed at 0.1. The best configuration for each benchmark is bolded.}
  \label{tab:segment4}
  \begin{tabular}{c|cc|cc|cc|cc}
    \toprule
    \multirow{3}{*}{\begin{tabular}[c]{@{}c@{}}Weight\\ Transfer\\ Method\end{tabular}} & \multicolumn{4}{c|}{\instancesA}                                                                                                   & \multicolumn{4}{c}{\instancesB}                                                                                                     \\ 
                                                                                             & \multicolumn{2}{c|}{\pmbest}                                           & \multicolumn{2}{c|}{\pmwrand}                     & \multicolumn{2}{c|}{\pmbest}                                            & \multicolumn{2}{c}{\pmwrand}                      \\ 
                                                                                             & {~\#solved~} & {~PAR-2~}   & {~\#solved~} & ~PAR-2~   & {~\#solved~} & {~PAR-2~}    & {~\#solved~} & ~PAR-2~    \\ \midrule
    \textbf{\intwt}                                                                               & {24}                & {23871}          & {25}                & 22538          & {98}                & {9223}          & {103}               & 9188          \\ \midrule
    \textbf{\linorigs{}}                                                                              & {26}                & {22256}          & {27}                & 21769          & {98}                & {9237}          & {104}               & 9189          \\ 
    \textbf{\lineqs}                                                                              & {\textbf{28}}       & {\textbf{21233}} & {27}       & 22228 & {111}      & {9129} & {113}      & 9114 \\ 
    \textbf{\linrevs}                                                                              & {26}       & {22142} & {\textbf{28}}       & 21338 & {115}      & {9082} & {\textbf{118}}      & \textbf{9055}
    \\  \bottomrule
  \end{tabular}
\end{table}

Scanning down the columns of Table~\ref{tab:segment4}, we see that all three linear weight configurations perform at least as well as the fixed weight version, regardless of variable selection method.
The improvements on the \instancesA{} benchmark are modest, with at most 4 additional solved instances.
The improvements on the \instancesB{} benchmark are more substantial, with a maximum of 17 additional solved instances.

Overall, the best-performing linear weight configuration was \linrevs{}, which transfers the more weight from clauses with the initial weight.
These results support prior findings that more weight should be freed up to the falsified clauses in local minima.
The best-performing variable selection method continues to be the weighted random method \pmwrand{}.


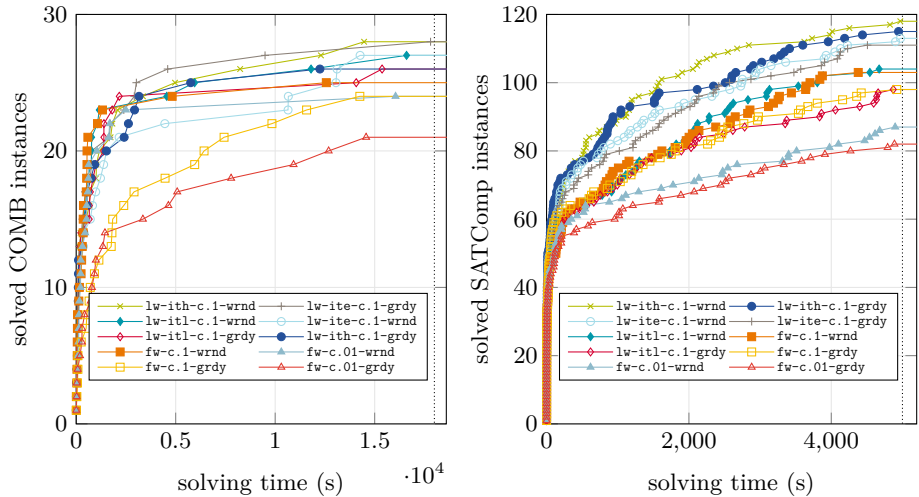
\begin{figure}
  \centering
      \begin{tikzpicture}[scale = 1]
          \begin{axis}[mark size=1.5pt, grid=both, grid style={black!10}, legend columns=2, legend style={nodes={scale=0.45]}, at={(0.95,0.32)}}, legend cell align={left}, 
                              ylabel near ticks, ylabel near ticks, ylabel shift={0pt}, width=185pt, height=200pt,xlabel=solving time (s), ylabel=solved COMB instances,xmin=0,xmax=18600,ymin=0,ymax=30]
          
\addplot[color=colh, mark=\markh] coordinates {(4.61, 1) (5.87, 2) (7.68, 3) (8.40, 4) (24.89, 5) (28.65, 6) (51.13, 7) (56.21, 8) (90.49, 9) (109.80, 10) (114.66, 11) (116.58, 12) (176.78, 13) (223.66, 14) (495.86, 15) (656.47, 16) (692.82, 17) (1001.36, 18) (1015.94, 19) (1164.79, 20) (1726.70, 21) (1818.43, 22) (2077.01, 23) (3353.45, 24) (4966.84, 25) (8239.66, 26) (12299.34, 27) (14467.61, 28) (36000, 28) };
\addplot[color=cole, mark=\marke] coordinates {(5.36, 1) (8.88, 2) (11.94, 3) (31.60, 4) (51.32, 5) (56.68, 6) (79.20, 7) (80.02, 8) (84.20, 9) (85.20, 10) (90.76, 11) (122.21, 12) (148.00, 13) (269.00, 14) (313.76, 15) (410.25, 16) (440.35, 17) (716.51, 18) (761.22, 19) (773.12, 20) (1660.01, 21) (1721.97, 22) (2393.01, 23) (2920.33, 24) (3024.64, 25) (4597.72, 26) (9490.12, 27) (17807.83, 28) (36000, 28) };
\addplot[color=colj, mark=\markj] coordinates {(7.73, 1) (11.23, 2) (12.76, 3) (33.69, 4) (34.12, 5) (80.70, 6) (86.94, 7) (98.17, 8) (128.67, 9) (156.86, 10) (178.98, 11) (208.82, 12) (264.88, 13) (332.85, 14) (372.25, 15) (465.77, 16) (517.88, 17) (563.84, 18) (757.53, 19) (770.71, 20) (775.88, 21) (1070.57, 22) (1158.95, 23) (4551.09, 24) (5952.52, 25) (11802.80, 26) (16602.39, 27) (36000, 27) };
\addplot[color=colf, mark=\markf] coordinates {(7.42, 1) (16.61, 2) (31.35, 3) (50.58, 4) (69.45, 5) (73.21, 6) (82.82, 7) (108.13, 8) (119.58, 9) (136.54, 10) (143.82, 11) (267.72, 12) (362.52, 13) (392.48, 14) (707.88, 15) (820.84, 16) (979.33, 17) (1221.08, 18) (1378.28, 19) (1534.85, 20) (2158.41, 21) (4454.84, 22) (10646.33, 23) (10661.16, 24) (13054.93, 25) (13106.77, 26) (14267.43, 27) (36000, 27) };
\addplot[color=coli, mark=\marki] coordinates {(2.67, 1) (5.72, 2) (11.64, 3) (13.24, 4) (24.71, 5) (35.82, 6) (74.83, 7) (96.26, 8) (109.42, 9) (110.30, 10) (115.24, 11) (188.16, 12) (193.50, 13) (234.93, 14) (655.32, 15) (663.36, 16) (714.63, 17) (854.96, 18) (985.69, 19) (1380.87, 20) (1391.01, 21) (1394.27, 22) (1807.87, 23) (2164.84, 24) (14072.84, 25) (15372.54, 26) (36000, 26) };
\addplot[color=colg, mark=\markg] coordinates {(5.65, 1) (9.47, 2) (12.94, 3) (39.58, 4) (58.62, 5) (69.53, 6) (74.12, 7) (87.87, 8) (106.89, 9) (107.58, 10) (112.65, 11) (116.12, 12) (296.93, 13) (369.25, 14) (393.92, 15) (539.81, 16) (563.02, 17) (743.96, 18) (923.44, 19) (1515.97, 20) (2410.06, 21) (2624.54, 22) (2931.96, 23) (3156.08, 24) (5753.32, 25) (12247.79, 26) (36000, 26) };
\addplot[color=cold, mark=\markd] coordinates {(6.74, 1) (7.64, 2) (31.10, 3) (37.56, 4) (52.19, 5) (52.89, 6) (61.63, 7) (68.02, 8) (132.48, 9) (160.63, 10) (245.26, 11) (273.19, 12) (303.19, 13) (349.82, 14) (352.87, 15) (366.61, 16) (507.45, 17) (526.64, 18) (542.27, 19) (544.29, 20) (587.58, 21) (1070.39, 22) (1321.77, 23) (4822.37, 24) (12569.52, 25) (36000, 25) };
\addplot[color=colb, mark=\markb] coordinates {(0.99, 1) (4.79, 2) (17.88, 3) (81.96, 4) (86.32, 5) (119.12, 6) (135.14, 7) (141.44, 8) (160.96, 9) (216.72, 10) (223.40, 11) (230.96, 12) (314.73, 13) (445.75, 14) (517.87, 15) (524.12, 16) (586.63, 17) (618.19, 18) (631.97, 19) (1038.52, 20) (1637.98, 21) (1681.92, 22) (2494.69, 23) (16049.60, 24) (36000, 24) };
\addplot[color=colc, mark=\markc] coordinates {(6.96, 1) (34.29, 2) (62.91, 3) (88.79, 4) (90.25, 5) (253.19, 6) (500.32, 7) (534.18, 8) (647.74, 9) (708.28, 10) (916.59, 11) (1140.15, 12) (1760.07, 13) (1795.39, 14) (1833.65, 15) (2389.20, 16) (2906.86, 17) (4467.09, 18) (5954.80, 19) (6426.03, 20) (7431.48, 21) (9836.71, 22) (11577.28, 23) (14254.05, 24) (36000, 24) };
\addplot[color=cola, mark=\marka] coordinates {(4.84, 1) (22.59, 2) (29.72, 3) (77.10, 4) (191.83, 5) (287.45, 6) (295.52, 7) (422.01, 8) (729.11, 9) (789.26, 10) (917.02, 11) (989.81, 12) (1341.76, 13) (1445.36, 14) (3351.97, 15) (4632.40, 16) (5076.29, 17) (7784.57, 18) (10923.05, 19) (12686.75, 20) (14568.75, 21) (36000, 21) };

\legend{ \solverh, \solvere, \solverj, \solverf, \solveri, \solverg, \solverd, \solverb,\solverc,  \solvera }

\addplot[densely dotted, color=black] coordinates {(18000, -1) (18000, 400)};

          \end{axis}
        \end{tikzpicture}~
  \begin{tikzpicture}[scale = 1]
          \begin{axis}[mark size=1.5pt, grid=both, grid style={black!10}, legend columns=2, legend style={nodes={scale=0.45]}, at={(0.95,0.32)}}, legend cell align={left}, 
                              ylabel near ticks, ylabel near ticks, ylabel shift={-3pt}, width=185pt, height=200pt,xlabel=solving time (s), ylabel=solved SATComp instances,xmin=0,xmax=5200,ymin=0,ymax=120]
          
\addplot[color=colh, mark=\markh] coordinates {(0.01, 1) (0.01, 2) (0.01, 3) (0.02, 4) (0.04, 5) (0.04, 6) (0.05, 7) (0.07, 8) (0.08, 9) (0.19, 10) (0.23, 11) (0.25, 12) (0.31, 13) (0.35, 14) (0.45, 15) (0.48, 16) (0.51, 17) (0.79, 18) (0.92, 19) (0.99, 20) (1.18, 21) (1.21, 22) (1.31, 23) (1.32, 24) (1.40, 25) (1.94, 26) (4.40, 27) (4.43, 28) (5.06, 29) (5.88, 30) (6.02, 31) (6.24, 32) (6.32, 33) (7.47, 34) (8.03, 35) (8.30, 36) (8.47, 37) (8.60, 38) (10.84, 39) (13.26, 40) (13.26, 41) (14.15, 42) (14.38, 43) (15.08, 44) (17.73, 45) (17.74, 46) (18.38, 47) (23.22, 48) (24.72, 49) (27.02, 50) (32.18, 51) (33.46, 52) (41.57, 53) (42.06, 54) (44.04, 55) (47.17, 56) (51.32, 57) (58.31, 58) (63.66, 59) (78.34, 60) (79.07, 61) (85.97, 62) (86.81, 63) (98.16, 64) (170.57, 65) (177.39, 66) (204.18, 67) (208.98, 68) (210.14, 69) (223.49, 70) (226.76, 71) (265.01, 72) (284.24, 73) (308.47, 74) (339.84, 75) (373.14, 76) (400.23, 77) (501.42, 78) (517.85, 79) (522.34, 80) (551.88, 81) (552.44, 82) (556.14, 83) (599.23, 84) (724.09, 85) (741.93, 86) (846.39, 87) (958.83, 88) (982.06, 89) (1113.07, 90) (1121.82, 91) (1158.34, 92) (1162.27, 93) (1193.34, 94) (1198.06, 95) (1294.83, 96) (1478.75, 97) (1533.62, 98) (1557.55, 99) (1577.43, 100) (1615.68, 101) (1847.02, 102) (1904.54, 103) (2056.37, 104) (2116.94, 105) (2140.65, 106) (2252.07, 107) (2351.26, 108) (2547.71, 109) (2677.35, 110) (2841.90, 111) (3621.58, 112) (3730.65, 113) (3978.58, 114) (4012.30, 115) (4254.43, 116) (4824.59, 117) (4970.29, 118) (10000, 118) };
\addplot[color=colg, mark=\markg] coordinates {(0.01, 1) (0.02, 2) (0.03, 3) (0.03, 4) (0.05, 5) (0.06, 6) (0.07, 7) (0.13, 8) (0.16, 9) (0.29, 10) (0.31, 11) (0.48, 12) (0.49, 13) (0.53, 14) (0.53, 15) (0.81, 16) (0.97, 17) (0.99, 18) (1.08, 19) (1.21, 20) (1.40, 21) (1.57, 22) (1.98, 23) (2.40, 24) (3.36, 25) (3.46, 26) (4.11, 27) (4.66, 28) (5.66, 29) (5.93, 30) (5.94, 31) (5.94, 32) (6.91, 33) (6.98, 34) (7.79, 35) (8.22, 36) (8.24, 37) (8.90, 38) (9.25, 39) (9.63, 40) (10.94, 41) (12.71, 42) (13.24, 43) (13.48, 44) (13.69, 45) (13.95, 46) (14.78, 47) (14.92, 48) (14.97, 49) (17.64, 50) (29.57, 51) (39.40, 52) (41.86, 53) (47.31, 54) (51.42, 55) (52.78, 56) (54.87, 57) (58.33, 58) (59.27, 59) (76.37, 60) (83.37, 61) (89.93, 62) (91.83, 63) (96.52, 64) (100.27, 65) (117.01, 66) (139.48, 67) (141.34, 68) (158.83, 69) (158.96, 70) (186.10, 71) (199.82, 72) (290.55, 73) (339.57, 74) (346.92, 75) (394.51, 76) (494.98, 77) (589.66, 78) (652.55, 79) (704.32, 80) (721.38, 81) (748.45, 82) (782.33, 83) (814.75, 84) (850.16, 85) (853.70, 86) (865.92, 87) (900.11, 88) (927.43, 89) (934.34, 90) (1042.11, 91) (1045.30, 92) (1170.14, 93) (1514.48, 94) (1515.61, 95) (1548.12, 96) (1581.56, 97) (2116.56, 98) (2345.15, 99) (2511.65, 100) (2596.52, 101) (2604.39, 102) (2658.08, 103) (2862.07, 104) (2906.12, 105) (3022.85, 106) (3192.89, 107) (3287.91, 108) (3340.70, 109) (3413.50, 110) (3599.99, 111) (3957.53, 112) (4187.53, 113) (4438.32, 114) (4945.24, 115) (10000, 115) };
\addplot[color=colf, mark=\markf] coordinates {(0.00, 1) (0.01, 2) (0.02, 3) (0.03, 4) (0.04, 5) (0.06, 6) (0.08, 7) (0.14, 8) (0.20, 9) (0.20, 10) (0.20, 11) (0.27, 12) (0.41, 13) (0.46, 14) (0.48, 15) (0.63, 16) (0.75, 17) (0.78, 18) (0.80, 19) (1.09, 20) (1.11, 21) (1.16, 22) (1.16, 23) (1.22, 24) (2.67, 25) (3.37, 26) (3.42, 27) (3.50, 28) (4.22, 29) (4.95, 30) (5.14, 31) (6.78, 32) (7.76, 33) (8.56, 34) (10.13, 35) (12.43, 36) (12.75, 37) (12.99, 38) (13.50, 39) (14.48, 40) (14.57, 41) (15.79, 42) (17.66, 43) (18.24, 44) (22.24, 45) (23.29, 46) (24.17, 47) (31.99, 48) (35.03, 49) (56.30, 50) (63.49, 51) (84.01, 52) (90.12, 53) (98.85, 54) (100.29, 55) (119.19, 56) (120.02, 57) (127.27, 58) (131.20, 59) (139.06, 60) (150.14, 61) (151.30, 62) (170.35, 63) (179.29, 64) (183.28, 65) (185.94, 66) (188.23, 67) (194.30, 68) (218.42, 69) (271.06, 70) (276.95, 71) (338.29, 72) (354.25, 73) (429.96, 74) (439.03, 75) (478.69, 76) (519.59, 77) (707.50, 78) (727.25, 79) (769.22, 80) (804.02, 81) (895.61, 82) (1000.45, 83) (1089.46, 84) (1152.27, 85) (1181.59, 86) (1275.87, 87) (1337.03, 88) (1366.73, 89) (1484.91, 90) (1519.89, 91) (1585.04, 92) (1860.87, 93) (1931.72, 94) (2086.87, 95) (2322.26, 96) (2363.30, 97) (2573.58, 98) (2727.45, 99) (2741.66, 100) (2751.82, 101) (2924.71, 102) (2941.38, 103) (3000.27, 104) (3153.48, 105) (3358.70, 106) (3899.28, 107) (3927.30, 108) (4050.48, 109) (4129.51, 110) (4227.94, 111) (4902.36, 112) (4972.03, 113) (10000, 113) };
\addplot[color=cole, mark=\marke] coordinates {(0.01, 1) (0.02, 2) (0.03, 3) (0.04, 4) (0.05, 5) (0.06, 6) (0.07, 7) (0.08, 8) (0.08, 9) (0.09, 10) (0.24, 11) (0.46, 12) (0.49, 13) (0.69, 14) (0.73, 15) (0.82, 16) (0.87, 17) (1.12, 18) (1.17, 19) (1.25, 20) (1.48, 21) (1.86, 22) (1.89, 23) (2.16, 24) (2.39, 25) (2.40, 26) (3.19, 27) (3.71, 28) (3.82, 29) (3.84, 30) (3.85, 31) (5.06, 32) (5.67, 33) (5.80, 34) (6.97, 35) (7.79, 36) (9.71, 37) (11.23, 38) (11.85, 39) (13.74, 40) (14.46, 41) (14.98, 42) (15.34, 43) (15.85, 44) (17.70, 45) (18.93, 46) (22.34, 47) (22.74, 48) (25.17, 49) (28.62, 50) (39.32, 51) (41.15, 52) (48.18, 53) (50.16, 54) (74.69, 55) (80.38, 56) (108.95, 57) (114.81, 58) (118.83, 59) (138.95, 60) (140.07, 61) (146.58, 62) (153.73, 63) (176.42, 64) (213.41, 65) (224.11, 66) (250.69, 67) (269.54, 68) (381.71, 69) (381.79, 70) (427.92, 71) (449.19, 72) (585.80, 73) (587.08, 74) (633.40, 75) (746.62, 76) (794.29, 77) (811.51, 78) (839.95, 79) (1022.41, 80) (1213.94, 81) (1215.55, 82) (1249.06, 83) (1317.12, 84) (1402.87, 85) (1492.41, 86) (1567.89, 87) (1584.55, 88) (1648.27, 89) (1708.52, 90) (1810.82, 91) (1892.64, 92) (2015.44, 93) (2080.39, 94) (2108.63, 95) (2138.88, 96) (2341.25, 97) (2380.28, 98) (2418.33, 99) (2570.23, 100) (2871.23, 101) (3152.16, 102) (3474.88, 103) (3570.72, 104) (3859.74, 105) (3905.98, 106) (4084.56, 107) (4117.52, 108) (4129.27, 109) (4229.70, 110) (4508.07, 111) (10000, 111) };
\addplot[color=colj, mark=\markj] coordinates {(0.01, 1) (0.02, 2) (0.03, 3) (0.04, 4) (0.05, 5) (0.07, 6) (0.07, 7) (0.07, 8) (0.08, 9) (0.11, 10) (0.12, 11) (0.17, 12) (0.19, 13) (0.27, 14) (0.40, 15) (0.43, 16) (0.47, 17) (0.58, 18) (1.05, 19) (1.07, 20) (1.16, 21) (1.39, 22) (1.62, 23) (1.82, 24) (1.89, 25) (2.61, 26) (3.14, 27) (3.36, 28) (6.49, 29) (7.58, 30) (10.75, 31) (11.50, 32) (12.39, 33) (17.18, 34) (17.66, 35) (20.38, 36) (21.55, 37) (23.21, 38) (25.36, 39) (31.70, 40) (32.44, 41) (36.87, 42) (38.50, 43) (40.07, 44) (43.15, 45) (43.93, 46) (49.63, 47) (51.50, 48) (63.02, 49) (74.72, 50) (76.21, 51) (80.43, 52) (112.01, 53) (117.80, 54) (145.07, 55) (145.27, 56) (158.92, 57) (161.28, 58) (171.77, 59) (186.61, 60) (266.98, 61) (299.00, 62) (344.82, 63) (473.76, 64) (527.81, 65) (604.93, 66) (780.69, 67) (913.26, 68) (919.22, 69) (926.90, 70) (1010.13, 71) (1061.43, 72) (1113.63, 73) (1210.20, 74) (1263.58, 75) (1269.29, 76) (1303.35, 77) (1484.07, 78) (1725.91, 79) (1737.20, 80) (1789.55, 81) (1821.56, 82) (1931.58, 83) (2006.22, 84) (2062.71, 85) (2066.92, 86) (2100.59, 87) (2122.43, 88) (2279.70, 89) (2284.71, 90) (2502.93, 91) (2544.03, 92) (2579.97, 93) (2623.48, 94) (2829.56, 95) (3089.12, 96) (3093.78, 97) (3131.14, 98) (3536.94, 99) (3804.00, 100) (3832.00, 101) (3886.54, 102) (4542.28, 103) (4672.54, 104) (10000, 104) };
\addplot[color=cold, mark=\markd] coordinates {(0.01, 1) (0.01, 2) (0.01, 3) (0.02, 4) (0.10, 5) (0.10, 6) (0.11, 7) (0.14, 8) (0.14, 9) (0.14, 10) (0.23, 11) (0.24, 12) (0.30, 13) (0.33, 14) (0.38, 15) (0.39, 16) (0.44, 17) (0.45, 18) (0.47, 19) (0.66, 20) (0.76, 21) (0.88, 22) (1.09, 23) (1.16, 24) (1.60, 25) (1.76, 26) (2.04, 27) (2.12, 28) (2.97, 29) (5.42, 30) (5.49, 31) (6.76, 32) (6.91, 33) (7.45, 34) (14.31, 35) (16.77, 36) (17.47, 37) (18.47, 38) (19.76, 39) (21.27, 40) (21.59, 41) (22.50, 42) (26.49, 43) (28.23, 44) (33.03, 45) (38.38, 46) (51.44, 47) (51.44, 48) (68.88, 49) (130.86, 50) (130.93, 51) (137.87, 52) (151.67, 53) (151.93, 54) (177.82, 55) (198.98, 56) (207.75, 57) (209.64, 58) (227.02, 59) (260.81, 60) (285.33, 61) (290.34, 62) (353.31, 63) (469.72, 64) (471.20, 65) (614.52, 66) (658.64, 67) (777.72, 68) (797.66, 69) (821.83, 70) (840.62, 71) (888.04, 72) (919.42, 73) (935.32, 74) (1009.40, 75) (1125.31, 76) (1158.80, 77) (1482.07, 78) (1562.37, 79) (1620.15, 80) (1805.79, 81) (1952.16, 82) (1961.83, 83) (2002.50, 84) (2048.27, 85) (2140.61, 86) (2376.75, 87) (2579.71, 88) (2580.24, 89) (2650.63, 90) (2825.22, 91) (2857.34, 92) (3117.31, 93) (3187.67, 94) (3252.78, 95) (3271.02, 96) (3439.74, 97) (3640.22, 98) (3644.11, 99) (3701.17, 100) (3839.25, 101) (3874.38, 102) (4378.93, 103) (10000, 103) };
\addplot[color=coli, mark=\marki] coordinates {(0.01, 1) (0.01, 2) (0.01, 3) (0.03, 4) (0.03, 5) (0.03, 6) (0.08, 7) (0.10, 8) (0.11, 9) (0.22, 10) (0.24, 11) (0.26, 12) (0.27, 13) (0.30, 14) (0.32, 15) (0.47, 16) (0.52, 17) (0.53, 18) (0.68, 19) (0.85, 20) (0.88, 21) (1.04, 22) (1.17, 23) (1.39, 24) (1.71, 25) (2.16, 26) (2.31, 27) (3.34, 28) (5.74, 29) (8.48, 30) (9.43, 31) (12.43, 32) (12.66, 33) (13.76, 34) (15.99, 35) (18.93, 36) (27.83, 37) (28.52, 38) (33.23, 39) (33.33, 40) (33.36, 41) (43.92, 42) (44.53, 43) (45.28, 44) (50.26, 45) (56.66, 46) (74.54, 47) (75.50, 48) (79.12, 49) (105.47, 50) (108.44, 51) (118.74, 52) (127.51, 53) (140.19, 54) (144.37, 55) (145.53, 56) (163.76, 57) (184.21, 58) (226.70, 59) (247.02, 60) (298.45, 61) (420.67, 62) (499.88, 63) (564.03, 64) (663.70, 65) (698.39, 66) (753.13, 67) (856.28, 68) (928.95, 69) (993.67, 70) (1025.94, 71) (1038.83, 72) (1162.71, 73) (1233.28, 74) (1261.88, 75) (1328.55, 76) (1391.96, 77) (1473.37, 78) (1744.41, 79) (1892.84, 80) (2030.03, 81) (2077.46, 82) (2090.50, 83) (2147.87, 84) (2486.95, 85) (2617.04, 86) (2782.98, 87) (3352.10, 88) (3373.45, 89) (3836.13, 90) (3862.42, 91) (4058.47, 92) (4399.20, 93) (4595.73, 94) (4634.55, 95) (4637.99, 96) (4658.42, 97) (4870.98, 98) (10000, 98) };
\addplot[color=colc, mark=\markc] coordinates {(0.02, 1) (0.03, 2) (0.03, 3) (0.03, 4) (0.03, 5) (0.04, 6) (0.05, 7) (0.07, 8) (0.16, 9) (0.19, 10) (0.23, 11) (0.26, 12) (0.31, 13) (0.32, 14) (0.40, 15) (0.43, 16) (0.54, 17) (0.63, 18) (0.77, 19) (0.90, 20) (1.22, 21) (1.42, 22) (1.46, 23) (1.60, 24) (1.86, 25) (2.43, 26) (2.88, 27) (4.42, 28) (5.44, 29) (6.70, 30) (6.79, 31) (7.02, 32) (8.50, 33) (11.22, 34) (13.90, 35) (15.35, 36) (19.12, 37) (20.57, 38) (21.36, 39) (25.23, 40) (28.76, 41) (28.84, 42) (32.33, 43) (32.96, 44) (33.92, 45) (33.92, 46) (36.32, 47) (46.71, 48) (50.82, 49) (55.12, 50) (55.69, 51) (58.41, 52) (62.09, 53) (64.21, 54) (71.72, 55) (74.56, 56) (85.53, 57) (100.83, 58) (103.63, 59) (128.89, 60) (144.06, 61) (211.69, 62) (269.36, 63) (331.12, 64) (516.94, 65) (596.92, 66) (631.29, 67) (662.43, 68) (855.04, 69) (890.17, 70) (1035.45, 71) (1077.27, 72) (1186.40, 73) (1196.67, 74) (1283.22, 75) (1355.33, 76) (1452.25, 77) (1595.72, 78) (1751.81, 79) (1807.99, 80) (1855.91, 81) (2262.21, 82) (2305.19, 83) (2429.43, 84) (2461.38, 85) (2476.73, 86) (2546.66, 87) (2771.55, 88) (2962.13, 89) (2977.15, 90) (3538.69, 91) (3567.68, 92) (3670.23, 93) (3803.80, 94) (4158.99, 95) (4356.45, 96) (4420.87, 97) (4919.98, 98) (10000, 98) };
\addplot[color=colb, mark=\markb] coordinates {(0.00, 1) (0.01, 2) (0.01, 3) (0.02, 4) (0.03, 5) (0.05, 6) (0.06, 7) (0.06, 8) (0.06, 9) (0.06, 10) (0.07, 11) (0.09, 12) (0.30, 13) (0.41, 14) (0.42, 15) (1.00, 16) (1.22, 17) (1.32, 18) (1.70, 19) (1.74, 20) (1.81, 21) (1.87, 22) (2.50, 23) (3.08, 24) (3.76, 25) (4.11, 26) (4.23, 27) (4.23, 28) (4.49, 29) (5.30, 30) (5.75, 31) (6.81, 32) (8.63, 33) (12.74, 34) (13.39, 35) (15.71, 36) (16.22, 37) (17.94, 38) (23.86, 39) (27.37, 40) (27.96, 41) (32.39, 42) (37.71, 43) (39.92, 44) (42.00, 45) (65.98, 46) (88.96, 47) (95.31, 48) (99.09, 49) (113.44, 50) (119.50, 51) (121.87, 52) (132.09, 53) (162.38, 54) (165.66, 55) (172.48, 56) (208.47, 57) (209.86, 58) (305.84, 59) (348.13, 60) (428.88, 61) (520.60, 62) (540.81, 63) (547.35, 64) (880.85, 65) (1056.60, 66) (1112.29, 67) (1293.05, 68) (1609.98, 69) (1776.85, 70) (2073.70, 71) (2138.80, 72) (2463.49, 73) (2474.65, 74) (2545.77, 75) (2677.15, 76) (3315.09, 77) (3393.32, 78) (3400.88, 79) (3512.58, 80) (3975.29, 81) (4100.79, 82) (4468.01, 83) (4537.37, 84) (4615.36, 85) (4679.95, 86) (4897.45, 87) (10000, 87) };
\addplot[color=cola, mark=\marka] coordinates {(0.01, 1) (0.02, 2) (0.03, 3) (0.03, 4) (0.11, 5) (0.11, 6) (0.12, 7) (0.12, 8) (0.13, 9) (0.14, 10) (0.14, 11) (0.16, 12) (0.22, 13) (0.30, 14) (0.32, 15) (0.40, 16) (0.60, 17) (0.62, 18) (0.96, 19) (1.16, 20) (1.69, 21) (2.14, 22) (3.00, 23) (3.66, 24) (5.25, 25) (5.29, 26) (5.30, 27) (5.64, 28) (5.89, 29) (6.11, 30) (6.19, 31) (7.37, 32) (8.89, 33) (10.81, 34) (11.44, 35) (14.02, 36) (18.33, 37) (20.31, 38) (23.20, 39) (28.85, 40) (41.58, 41) (42.76, 42) (50.60, 43) (59.92, 44) (83.82, 45) (99.69, 46) (103.93, 47) (108.63, 48) (116.79, 49) (119.86, 50) (133.45, 51) (167.09, 52) (205.12, 53) (205.69, 54) (209.98, 55) (385.20, 56) (418.38, 57) (581.46, 58) (641.58, 59) (956.30, 60) (1018.24, 61) (1018.45, 62) (1071.72, 63) (1211.54, 64) (1577.73, 65) (1584.51, 66) (1887.19, 67) (2073.65, 68) (2198.92, 69) (2445.22, 70) (2515.09, 71) (2562.47, 72) (2870.96, 73) (3005.72, 74) (3064.21, 75) (3332.68, 76) (3583.45, 77) (3707.54, 78) (3910.77, 79) (4269.77, 80) (4792.06, 81) (4887.74, 82) (10000, 82) };

\legend{ \solverh, \solverg, \solverf, \solvere, \solverj, \solverd, \solveri, \solverc, \solverb, \solvera }

\addplot[densely dotted, color=black] coordinates {(5000, -1) (5000, 400)};

          \end{axis}
        \end{tikzpicture}
  
  \caption{Performance profiles of \yalsatddfw{} (\texttt{fw-c.01-grdy}) and nine modifications for \instancesA{} (left) and \instancesB{} (right).}
  \label{fig:comb+sclarge}
\end{figure}
\vspace{-.6cm}
\noindent\textbf{Analysis of solve count over runtime.}
In addition to solve counts and PAR-2 scores for the three linear weight configurations, we report solve counts as a function of solving time.
The data for ten experimental settings of \yalsatddfw{} on the two benchmarks are shown in Figure~\ref{fig:comb+sclarge}.
Note that the original \ddfw{} setting is represented by the setting \texttt{\intwtshort-c.01-\pmbest}, and is our baseline.

For the \instancesA{} benchmark (Figure~\ref{fig:comb+sclarge}, left plot), all nine other settings (our modifications) outperform the baseline in terms of solving speed and number of solved instances.
The best settings are \texttt{\linrevs{}-c.1-\pmwrand} and \texttt{\lineqs{}-c.1-\pmbest}, which perform on par with each other and solve 28 instances by timeout.
For the \instancesB{} benchmark (Figure~\ref{fig:comb+sclarge}, right plot), the dominance of the setting \texttt{\linrevs{}-c.1-wrnd} is more pronounced.
For about the first 1,000 seconds, this setting performs similar to \texttt{\linrevs{}-c.1-\pmbest}.
After that, however, it begins to perform the best of all the settings, and it ends up solving the most instances by timeout, at 118.
The baseline setting \texttt{\intwtshort-c.01-\pmbest} ends up solving 83 instances at timeout, which is 35 less than \texttt{\linrevs{}-c.1-\pmwrand}. 

These two plots clearly show that our modifications substantially improve the original \ddfw{} algorithm.

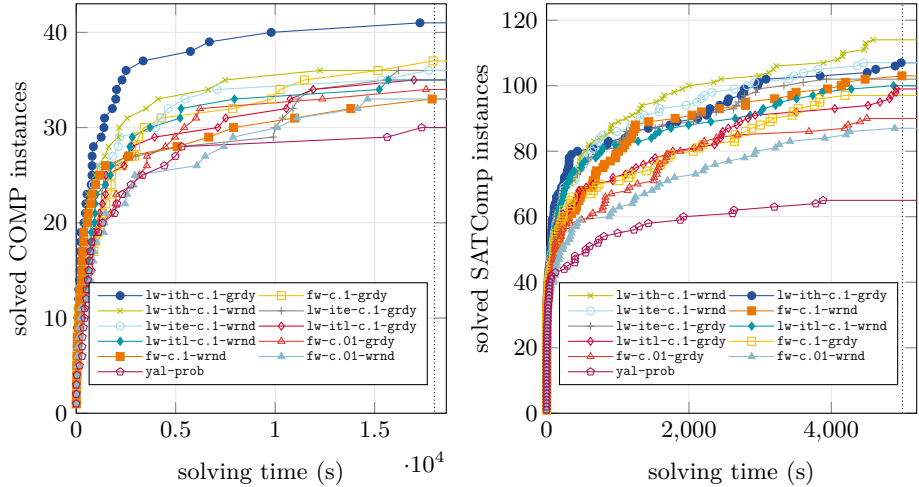
\begin{figure}
  \centering

    \begin{tikzpicture}[scale = 1]
          \begin{axis}[mark size=1.5pt, grid=both, grid style={black!10}, legend columns=2, legend style={nodes={scale=0.45]}, at={(0.95,0.32)}}, legend cell align={left}, 
                              ylabel near ticks, ylabel near ticks, ylabel shift={0pt}, width=185pt, height=200pt,xlabel=solving time (s), ylabel=solved COMP instances,xmin=0,xmax=18600,ymin=0,ymax=43]
          
\addplot[color=colg, mark=\markg] coordinates {(5.68, 1) (6.49, 2) (15.97, 3) (24.38, 4) (29.76, 5) (37.62, 6) (50.62, 7) (58.35, 8) (74.14, 9) (83.66, 10) (100.96, 11) (124.52, 12) (127.63, 13) (163.03, 14) (188.20, 15) (198.00, 16) (215.82, 17) (223.78, 18) (226.11, 19) (386.97, 20) (464.77, 21) (484.41, 22) (541.99, 23) (751.14, 24) (794.96, 25) (799.32, 26) (799.72, 27) (863.00, 28) (1256.88, 29) (1394.80, 30) (1424.13, 31) (1817.10, 32) (1988.05, 33) (2026.81, 34) (2304.54, 35) (2507.07, 36) (3364.15, 37) (5739.14, 38) (6691.53, 39) (9799.67, 40) (17275.72, 41) (36000, 41) };
\addplot[color=colc, mark=\markc] coordinates {(3.23, 1) (4.88, 2) (13.41, 3) (28.06, 4) (31.69, 5) (32.81, 6) (86.70, 7) (97.53, 8) (118.60, 9) (121.76, 10) (125.10, 11) (177.93, 12) (186.32, 13) (356.64, 14) (452.08, 15) (547.88, 16) (884.17, 17) (952.46, 18) (1099.37, 19) (1123.49, 20) (1328.92, 21) (1481.47, 22) (1746.76, 23) (1755.77, 24) (1768.10, 25) (2340.11, 26) (2941.40, 27) (2982.93, 28) (3138.35, 29) (3416.87, 30) (4301.37, 31) (7871.00, 32) (9828.26, 33) (10240.23, 34) (11467.47, 35) (15163.64, 36) (17935.23, 37) (36000, 37) };
\addplot[color=colh, mark=\markh] coordinates {(7.76, 1) (11.53, 2) (24.49, 3) (30.03, 4) (71.73, 5) (78.22, 6) (91.81, 7) (97.53, 8) (103.06, 9) (109.85, 10) (187.82, 11) (193.21, 12) (216.00, 13) (223.93, 14) (301.73, 15) (334.11, 16) (337.73, 17) (351.18, 18) (378.86, 19) (409.99, 20) (491.51, 21) (554.19, 22) (596.48, 23) (905.23, 24) (953.33, 25) (1132.82, 26) (1440.83, 27) (1623.94, 28) (2113.35, 29) (2161.14, 30) (2562.37, 31) (3516.11, 32) (4116.07, 33) (6637.94, 34) (7482.09, 35) (12252.15, 36) (36000, 36) };
\addplot[color=cole, mark=\marke] coordinates {(7.72, 1) (14.44, 2) (19.57, 3) (31.36, 4) (56.48, 5) (61.66, 6) (75.13, 7) (82.86, 8) (104.12, 9) (149.84, 10) (212.79, 11) (272.00, 12) (351.49, 13) (380.78, 14) (414.24, 15) (418.85, 16) (560.17, 17) (775.42, 18) (956.56, 19) (969.36, 20) (1207.62, 21) (1266.02, 22) (1292.27, 23) (1367.81, 24) (1417.71, 25) (1953.19, 26) (2996.92, 27) (5212.05, 28) (9906.27, 29) (9991.46, 30) (10366.05, 31) (10893.83, 32) (11153.39, 33) (11855.68, 34) (15585.31, 35) (16191.80, 36) (36000, 36) };
\addplot[color=colf, mark=\markf] coordinates {(5.49, 1) (9.20, 2) (11.95, 3) (15.29, 4) (35.88, 5) (109.38, 6) (120.43, 7) (168.09, 8) (216.39, 9) (257.67, 10) (330.74, 11) (330.91, 12) (406.31, 13) (410.25, 14) (421.53, 15) (576.95, 16) (586.58, 17) (633.82, 18) (709.72, 19) (742.68, 20) (756.83, 21) (824.65, 22) (851.07, 23) (857.29, 24) (1190.70, 25) (1292.93, 26) (1841.50, 27) (2115.42, 28) (2186.42, 29) (3994.57, 30) (4230.13, 31) (4617.35, 32) (5468.02, 33) (7074.89, 34) (15336.71, 35) (17712.66, 36) (36000, 36) };
\addplot[color=coli, mark=\marki] coordinates {(7.83, 1) (16.68, 2) (25.37, 3) (34.02, 4) (50.54, 5) (118.67, 6) (128.44, 7) (130.25, 8) (168.25, 9) (332.43, 10) (364.67, 11) (462.95, 12) (600.02, 13) (686.98, 14) (800.90, 15) (804.66, 16) (835.46, 17) (884.29, 18) (894.87, 19) (1157.53, 20) (1183.10, 21) (1320.37, 22) (1477.92, 23) (1479.60, 24) (1500.16, 25) (2408.85, 26) (2522.39, 27) (2789.53, 28) (3942.59, 29) (7112.75, 30) (7514.06, 31) (10562.16, 32) (10750.79, 33) (11912.79, 34) (16971.85, 35) (36000, 35) };
\addplot[color=colj, mark=\markj] coordinates {(1.08, 1) (5.83, 2) (9.02, 3) (11.71, 4) (63.10, 5) (84.35, 6) (86.34, 7) (96.19, 8) (96.92, 9) (179.93, 10) (228.93, 11) (287.89, 12) (415.38, 13) (508.19, 14) (598.78, 15) (615.49, 16) (708.45, 17) (740.82, 18) (752.25, 19) (905.46, 20) (948.33, 21) (956.96, 22) (1013.42, 23) (1467.67, 24) (1715.30, 25) (1743.84, 26) (2540.37, 27) (2697.26, 28) (2820.68, 29) (3712.06, 30) (5220.16, 31) (5271.33, 32) (7954.63, 33) (15230.30, 34) (15688.22, 35) (36000, 35) };
\addplot[color=cola, mark=\marka] coordinates {(4.87, 1) (5.00, 2) (28.14, 3) (33.97, 4) (61.08, 5) (81.58, 6) (83.83, 7) (114.28, 8) (191.04, 9) (222.76, 10) (295.90, 11) (394.30, 12) (542.75, 13) (545.22, 14) (595.96, 15) (663.38, 16) (748.17, 17) (903.06, 18) (1050.08, 19) (1362.07, 20) (1726.54, 21) (1969.61, 22) (2012.49, 23) (2853.55, 24) (3169.29, 25) (3551.20, 26) (3559.62, 27) (4483.73, 28) (4968.33, 29) (5432.67, 30) (6019.98, 31) (6222.50, 32) (12373.49, 33) (17578.43, 34) (36000, 34) };
\addplot[color=cold, mark=\markd] coordinates {(6.87, 1) (15.62, 2) (23.80, 3) (31.60, 4) (37.42, 5) (52.76, 6) (54.44, 7) (54.46, 8) (104.07, 9) (142.96, 10) (165.44, 11) (219.67, 12) (232.21, 13) (270.30, 14) (289.48, 15) (290.31, 16) (329.14, 17) (372.35, 18) (372.91, 19) (566.72, 20) (604.54, 21) (730.37, 22) (797.02, 23) (971.24, 24) (1179.15, 25) (1472.65, 26) (2634.29, 27) (5067.02, 28) (6662.95, 29) (7911.69, 30) (10979.83, 31) (13801.66, 32) (17886.66, 33) (36000, 33) };
\addplot[color=colb, mark=\markb] coordinates {(6.86, 1) (50.53, 2) (56.29, 3) (62.72, 4) (62.89, 5) (70.94, 6) (90.61, 7) (139.24, 8) (156.16, 9) (219.30, 10) (223.62, 11) (504.75, 12) (673.61, 13) (717.21, 14) (750.48, 15) (871.54, 16) (897.29, 17) (1012.21, 18) (1393.71, 19) (1421.17, 20) (1485.72, 21) (2458.60, 22) (2559.87, 23) (2871.43, 24) (2947.07, 25) (6057.92, 26) (6483.71, 27) (7425.72, 28) (7882.44, 29) (10107.60, 30) (11094.90, 31) (14145.22, 32) (14642.54, 33) (36000, 33) };
\addplot[color=colk, mark=\markk] coordinates {(0.84, 1) (1.85, 2) (3.12, 3) (36.21, 4) (173.97, 5) (296.03, 6) (298.02, 7) (309.99, 8) (364.90, 9) (419.52, 10) (423.55, 11) (511.51, 12) (521.15, 13) (583.47, 14) (628.23, 15) (660.27, 16) (701.08, 17) (767.76, 18) (1112.59, 19) (1328.78, 20) (1928.82, 21) (2053.12, 22) (2320.21, 23) (2746.15, 24) (3332.84, 25) (4432.80, 26) (4984.72, 27) (5310.30, 28) (15632.29, 29) (17360.46, 30) (36000, 30) };

\legend{ \solverg, \solverc, \solverh, \solvere, \solverf, \solveri, \solverj, \solvera, \solverd, \solverb, \solverk }

\addplot[densely dotted, color=black] coordinates {(18000, -1) (18000, 400)};

          \end{axis}
        \end{tikzpicture}~
  \begin{tikzpicture}[scale = 1]
          \begin{axis}[mark size=1.5pt, grid=both, grid style={black!10}, legend columns=2, legend style={nodes={scale=0.45]}, at={(0.95,0.32)}}, legend cell align={left}, 
                              ylabel near ticks, ylabel near ticks, ylabel shift={-3pt}, width=185pt, height=200pt,xlabel=solving time (s), ylabel=solved SATComp instances,xmin=0,xmax=5200,ymin=0,ymax=125]
          
\addplot[color=colh, mark=\markh] coordinates {(0.01,1) (0.01,2) (0.02,3) (0.03,4) (0.04,5) (0.04,6) (0.05,7) (0.07,8) (0.07,9) (0.13,10) (0.19,11) (0.21,12) (0.23,13) (0.25,14) (0.31,15) (0.35,16) (0.4,17) (0.6,18) (0.63,19) (0.63,20) (0.69,21) (0.82,22) (0.92,23) (1.18,24) (1.41,25) (1.53,26) (2.26,27) (4.43,28) (6.02,29) (6.44,30) (6.62,31) (7.3,32) (8.16,33) (11.86,34) (13.96,35) (14.14,36) (15.24,37) (15.4,38) (17.58,39) (17.75,40) (17.91,41) (20.35,42) (20.8,43) (24.59,44) (29.06,45) (30.09,46) (30.74,47) (33.14,48) (35.85,49) (41.33,50) (71.78,51) (74.39,52) (87.49,53) (95.13,54) (98.16,55) (104.0,56) (109.86,57) (120.07,58) (126.81,59) (128.2,60) (129.1,61) (131.04,62) (145.72,63) (154.34,64) (156.52,65) (171.45,66) (183.22,67) (255.76,68) (265.14,69) (344.94,70) (380.59,71) (381.01,72) (393.61,73) (424.22,74) (445.05,75) (464.55,76) (467.87,77) (476.06,78) (487.16,79) (546.82,80) (588.56,81) (592.55,82) (666.48,83) (737.56,84) (843.86,85) (900.74,86) (902.16,87) (914.02,88) (1008.33,89) (1116.83,90) (1201.94,91) (1292.81,92) (1316.47,93) (1338.66,94) (1542.64,95) (1627.92,96) (1651.42,97) (1795.59,98) (1908.59,99) (2001.49,100) (2348.32,101) (2450.89,102) (2982.29,103) (3121.62,104) (3188.67,105) (3229.87,106) (3889.94,107) (4086.15,108) (4162.32,109) (4162.7,110) (4460.24,111) (4481.25,112) (4483.48,113) (4586.53,114) (10000, 114) };
\addplot[color=colg, mark=\markg] coordinates {(0.006838,1) (0.013385,2) (0.025369,3) (0.032866,4) (0.046782,5) (0.05657,6) (0.065954,7) (0.125546,8) (0.164783,9) (0.247211,10) (0.305552,11) (0.512472,12) (0.516252,13) (0.533949,14) (0.534526,15) (0.567624,16) (0.807844,17) (0.825334,18) (0.9406,19) (1.035026,20) (1.203105,21) (1.275893,22) (1.974797,23) (2.454492,24) (3.481532,25) (4.171895,26) (4.295097,27) (5.915277,28) (6.476271,29) (7.477532,30) (7.574832,31) (8.979726,32) (8.993617,33) (9.257369,34) (9.340972,35) (13.662742,36) (14.8221,37) (15.991336,38) (16.550531,39) (18.414104,40) (22.738622,41) (24.617866,42) (28.87383,43) (33.227118,44) (39.567975,45) (45.532944,46) (51.852473,47) (51.946278,48) (55.265764,49) (56.354119,50) (58.151087,51) (67.032997,52) (70.923229,53) (71.803727,54) (72.084684,55) (72.379979,56) (76.242721,57) (76.880389,58) (77.600437,59) (101.333078,60) (102.0249,61) (107.799876,62) (108.262176,63) (122.101986,64) (137.892953,65) (139.025358,66) (167.553417,67) (204.621351,68) (210.535117,69) (239.27245,70) (250.763465,71) (251.616033,72) (282.189274,73) (291.78597,74) (305.989451,75) (315.267851,76) (316.913833,77) (316.913833,78) (351.026435,79) (434.496799,80) (654.233982,81) (846.006266,82) (868.995314,83) (1101.11185,84) (1154.12496,85) (1226.938573,86) (1429.714646,87) (1731.863865,88) (1832.912421,89) (2121.314756,90) (2249.869525,91) (2268.19754,92) (2295.073521,93) (2432.630954,94) (2543.014839,95) (2558.364019,96) (2736.101147,97) (2769.462167,98) (2778.351402,99) (2963.79885,100) (2980.672094,101) (3081.813884,102) (3843.774798,103) (4504.88133,104) (4624.878644,105) (4897.054274,106) (4981.491878,107) (10000, 107) };
\addplot[color=colf, mark=\markf] coordinates {(0.0,1) (0.01,2) (0.03,3) (0.03,4) (0.04,5) (0.04,6) (0.06,7) (0.08,8) (0.14,9) (0.14,10) (0.19,11) (0.2,12) (0.2,13) (0.35,14) (0.46,15) (0.47,16) (0.48,17) (0.6,18) (0.79,19) (0.9,20) (1.09,21) (1.16,22) (1.2,23) (1.46,24) (1.88,25) (2.11,26) (2.16,27) (2.68,28) (2.89,29) (2.93,30) (3.38,31) (3.51,32) (3.68,33) (4.68,34) (5.5,35) (6.96,36) (17.68,37) (18.4,38) (18.94,39) (27.94,40) (28.18,41) (37.31,42) (41.49,43) (48.34,44) (50.0,45) (52.01,46) (56.44,47) (60.77,48) (70.25,49) (85.65,50) (87.49,51) (91.03,52) (96.86,53) (105.37,54) (109.41,55) (121.18,56) (127.5,57) (132.7,58) (154.66,59) (156.08,60) (172.02,61) (191.31,62) (210.68,63) (211.27,64) (229.54,65) (229.85,66) (230.38,67) (240.32,68) (244.28,69) (297.24,70) (390.89,71) (408.65,72) (417.1,73) (438.45,74) (480.16,75) (521.46,76) (539.68,77) (550.51,78) (610.65,79) (634.52,80) (668.85,81) (704.42,82) (743.89,83) (745.29,84) (795.25,85) (988.4,86) (1174.75,87) (1198.21,88) (1303.1,89) (1314.86,90) (1373.12,91) (1577.41,92) (1588.64,93) (1961.83,94) (2113.25,95) (2171.91,96) (2214.18,97) (2323.03,98) (2634.77,99) (2819.61,100) (2925.7,101) (3407.8,102) (3476.0,103) (3608.98,104) (3815.36,105) (4361.0,106) (4453.88,107) (10000, 107) };
\addplot[color=cold, mark=\markd] coordinates {(0.01,1) (0.01,2) (0.02,3) (0.02,4) (0.1,5) (0.1,6) (0.11,7) (0.12,8) (0.13,9) (0.14,10) (0.14,11) (0.14,12) (0.23,13) (0.25,14) (0.33,15) (0.38,16) (0.39,17) (0.43,18) (0.47,19) (0.48,20) (0.9,21) (1.03,22) (1.22,23) (1.76,24) (1.92,25) (2.11,26) (2.62,27) (2.84,28) (3.62,29) (3.91,30) (4.37,31) (4.72,32) (5.98,33) (7.09,34) (9.02,35) (16.74,36) (18.11,37) (22.52,38) (28.09,39) (40.96,40) (49.35,41) (54.73,42) (55.0,43) (63.75,44) (83.05,45) (90.64,46) (92.21,47) (99.42,48) (106.03,49) (124.1,50) (128.77,51) (139.17,52) (162.68,53) (177.38,54) (186.09,55) (199.2,56) (220.45,57) (222.79,58) (270.96,59) (305.41,60) (352.02,61) (400.11,62) (417.79,63) (418.78,64) (455.56,65) (472.87,66) (484.59,67) (505.89,68) (637.71,69) (649.4,70) (655.45,71) (685.02,72) (687.74,73) (703.77,74) (793.07,75) (813.06,76) (895.08,77) (967.05,78) (974.26,79) (1010.74,80) (1066.73,81) (1098.42,82) (1128.18,83) (1177.76,84) (1197.36,85) (1219.72,86) (1239.5,87) (1248.0,88) (1435.16,89) (1806.17,90) (2043.18,91) (2174.87,92) (2427.28,93) (2795.74,94) (2795.99,95) (3204.75,96) (3446.46,97) (3504.38,98) (3946.84,99) (4140.25,100) (4201.41,101) (4474.75,102) (4992.77,103) (10000, 103) };
\addplot[color=cole, mark=\marke] coordinates {(0.02,1) (0.02,2) (0.04,3) (0.05,4) (0.06,5) (0.06,6) (0.06,7) (0.07,8) (0.08,9) (0.08,10) (0.09,11) (0.24,12) (0.45,13) (0.46,14) (0.53,15) (0.73,16) (0.82,17) (0.89,18) (1.12,19) (1.25,20) (1.57,21) (1.86,22) (2.52,23) (4.02,24) (4.22,25) (5.84,26) (6.04,27) (6.4,28) (6.86,29) (7.76,30) (8.34,31) (9.02,32) (10.26,33) (11.97,34) (13.59,35) (14.0,36) (15.26,37) (15.8,38) (17.67,39) (19.4,40) (20.28,41) (22.91,42) (23.45,43) (31.49,44) (31.72,45) (31.86,46) (48.31,47) (51.0,48) (52.24,49) (66.73,50) (80.06,51) (94.42,52) (96.32,53) (98.49,54) (99.97,55) (115.93,56) (117.89,57) (122.08,58) (127.39,59) (127.79,60) (133.66,61) (139.01,62) (166.06,63) (183.42,64) (200.51,65) (200.66,66) (220.18,67) (232.44,68) (274.64,69) (282.9,70) (355.37,71) (366.44,72) (434.18,73) (477.49,74) (477.93,75) (511.23,76) (579.97,77) (588.81,78) (694.13,79) (748.13,80) (945.21,81) (1018.04,82) (1019.65,83) (1078.12,84) (1116.61,85) (1124.99,86) (1525.71,87) (1630.57,88) (2064.06,89) (2098.51,90) (2138.58,91) (2384.21,92) (2625.43,93) (2661.12,94) (2699.22,95) (2775.05,96) (2782.32,97) (2938.51,98) (2962.85,99) (3334.74,100) (3605.83,101) (3930.55,102) 
(10000, 102) };
\addplot[color=colj, mark=\markj] coordinates {(0.01,1) (0.02,2) (0.02,3) (0.04,4) (0.05,5) (0.07,6) (0.07,7) (0.07,8) (0.08,9) (0.11,10) (0.12,11) (0.17,12) (0.19,13) (0.26,14) (0.27,15) (0.47,16) (0.51,17) (0.79,18) (0.81,19) (0.89,20) (1.16,21) (1.39,22) (1.56,23) (1.61,24) (1.75,25) (2.33,26) (2.92,27) (3.72,28) (3.92,29) (5.85,30) (6.37,31) (6.44,32) (6.51,33) (9.55,34) (10.26,35) (12.45,36) (17.57,37) (18.72,38) (19.71,39) (21.75,40) (27.55,41) (29.84,42) (36.58,43) (46.92,44) (48.61,45) (49.48,46) (52.31,47) (58.95,48) (64.93,49) (66.67,50) (70.99,51) (88.05,52) (99.5,53) (109.8,54) (111.53,55) (114.94,56) (117.25,57) (154.27,58) (158.75,59) (158.81,60) (168.16,61) (170.35,62) (183.13,63) (201.06,64) (222.53,65) (226.97,66) (243.31,67) (250.74,68) (275.39,69) (283.88,70) (321.49,71) (321.93,72) (346.73,73) (388.22,74) (489.85,75) (496.87,76) (535.07,77) (691.44,78) (733.09,79) (774.74,80) (792.17,81) (966.4,82) (1285.09,83) (1471.76,84) (1524.52,85) (1582.78,86) (1747.5,87) (1995.1,88) (2308.29,89) (2651.08,90) (2858.98,91) (3010.45,92) (3044.17,93) (3217.04,94) (3280.14,95) (3510.41,96) (3741.52,97) (3873.25,98) (4280.69,99) (4875.34,100) (10000, 100) };
\addplot[color=coli, mark=\marki] coordinates {(0.01,1) (0.01,2) (0.01,3) (0.02,4) (0.03,5) (0.03,6) (0.1,7) (0.12,8) (0.12,9) (0.13,10) (0.22,11) (0.24,12) (0.26,13) (0.27,14) (0.28,15) (0.3,16) (0.32,17) (0.38,18) (0.5,19) (0.53,20) (0.58,21) (0.72,22) (0.88,23) (1.23,24) (1.24,25) (1.71,26) (2.53,27) (3.96,28) (4.05,29) (4.33,30) (4.82,31) (6.38,32) (6.83,33) (13.6,34) (16.5,35) (18.21,36) (18.45,37) (25.11,38) (30.9,39) (33.78,40) (39.07,41) (43.94,42) (45.72,43) (47.14,44) (47.85,45) (48.85,46) (49.38,47) (101.4,48) (114.22,49) (116.42,50) (118.24,51) (133.8,52) (135.35,53) (137.42,54) (140.04,55) (145.73,56) (147.43,57) (157.8,58) (183.07,59) (223.02,60) (230.23,61) (312.34,62) (325.52,63) (384.43,64) (410.77,65) (424.3,66) (437.74,67) (453.0,68) (556.01,69) (655.28,70) (931.56,71) (932.04,72) (1109.52,73) (1199.13,74) (1255.38,75) (1359.72,76) (1478.59,77) (1526.06,78) (1700.58,79) (1773.92,80) (2072.5,81) (2360.1,82) (2391.73,83) (2444.34,84) (2457.74,85) (2475.34,86) (2552.96,87) (2673.43,88) (2718.88,89) (2830.67,90) (2930.52,91) (3505.78,92) (4049.28,93) (4516.26,94) (4684.52,95) (4817.52,96) (4842.15,97) (4908.55,98) (4915.44,99) (10000, 99) };
\addplot[color=colc, mark=\markc] coordinates {(0.012212,1) (0.014557,2) (0.016453,3) (0.027173,4) (0.032056,5) (0.032524,6) (0.039243,7) (0.066039,8) (0.186136,9) (0.247564,10) (0.25621,11) (0.308848,12) (0.321326,13) (0.428926,14) (0.481173,15) (0.573409,16) (0.774478,17) (0.90417,18) (1.426479,19) (1.45772,20) (1.787801,21) (1.860597,22) (2.423687,23) (2.486683,24) (2.888554,25) (3.045196,26) (3.218505,27) (3.311191,28) (3.332893,29) (5.406313,30) (6.168143,31) (10.480632,32) (13.981991,33) (17.054778,34) (18.575102,35) (24.730862,36) (28.718763,37) (32.491012,38) (32.931014,39) (32.933869,40) (36.241914,41) (38.062169,42) (51.903071,43) (64.963501,44) (68.666495,45) (76.955867,46) (93.002672,47) (98.153032,48) (99.62847,49) (104.657105,50) (106.477327,51) (107.112664,52) (121.673423,53) (129.081417,54) (133.211886,55) (168.750434,56) (187.54162,57) (201.064329,58) (212.968217,59) (228.229314,60) (234.911648,61) (238.807431,62) (329.941477,63) (341.444949,64) (341.444949,65) (371.086175,66) (620.366374,67) (657.437647,68) (697.253338,69) (734.372495,70) (1016.681878,71) (1150.933378,72) (1184.077075,73) (1400.374596,74) (1527.253671,75) (1599.25226,76) (1622.221651,77) (1707.241839,78) (1746.201573,79) (2060.651767,80) (2128.833044,81) (2198.452332,82) (2638.334497,83) (2643.626307,84) (2651.700384,85) (2795.501549,86) (2883.418845,87) (3003.409037,88) (3139.403662,89) (3178.093183,90) (3449.0198,91) (3472.062531,92) (3532.910505,93) (3761.416184,94) (3838.26848,95) (3839.697007,96) (4192.764677,97) (10000, 97) };
\addplot[color=cola, mark=\marka] coordinates {(0.01,1) (0.02,2) (0.03,3) (0.03,4) (0.08,5) (0.11,6) (0.11,7) (0.12,8) (0.13,9) (0.13,10) (0.14,11) (0.16,12) (0.22,13) (0.29,14) (0.3,15) (0.4,16) (0.42,17) (0.46,18) (0.48,19) (0.6,20) (0.81,21) (0.83,22) (1.04,23) (1.69,24) (2.18,25) (3.06,26) (3.08,27) (3.66,28) (5.23,29) (7.36,30) (8.25,31) (8.74,32) (10.02,33) (12.73,34) (15.26,35) (22.45,36) (28.76,37) (31.08,38) (32.46,39) (41.62,40) (49.79,41) (55.89,42) (81.36,43) (98.03,44) (101.1,45) (101.11,46) (102.22,47) (106.1,48) (119.41,49) (119.74,50) (121.66,51) (121.78,52) (179.77,53) (202.47,54) (205.44,55) (221.46,56) (281.75,57) (331.69,58) (496.43,59) (610.38,60) (651.11,61) (782.35,62) (791.35,63) (815.52,64) (824.99,65) (826.41,66) (909.79,67) (1136.86,68) (1165.83,69) (1276.13,70) (1506.1,71) (1512.31,72) (1547.82,73) (1548.67,74) (1573.35,75) (1591.68,76) (1659.38,77) (1682.36,78) (1688.86,79) (1878.93,80) (2145.72,81) (2176.3,82) (2216.44,83) (2385.07,84) (2683.51,85) (3687.19,86) (4190.13,87) (4237.04,88) (4403.95,89) (4499.06,90) (10000, 90) };
\addplot[color=colb, mark=\markb] coordinates {(0.0,1) (0.01,2) (0.01,3) (0.02,4) (0.02,5) (0.03,6) (0.05,7) (0.06,8) (0.06,9) (0.06,10) (0.06,11) (0.08,12) (0.09,13) (0.3,14) (0.36,15) (0.42,16) (0.43,17) (0.47,18) (0.47,19) (1.06,20) (1.22,21) (1.45,22) (1.69,23) (1.7,24) (1.76,25) (1.87,26) (2.06,27) (3.4,28) (6.75,29) (9.89,30) (13.63,31) (17.86,32) (20.66,33) (24.05,34) (26.8,35) (42.04,36) (50.45,37) (58.16,38) (61.56,39) (88.94,40) (93.58,41) (120.57,42) (120.81,43) (129.96,44) (139.35,45) (145.98,46) (174.08,47) (181.87,48) (184.37,49) (231.11,50) (235.72,51) (284.32,52) (324.97,53) (331.62,54) (349.3,55) (350.09,56) (416.34,57) (427.82,58) (474.02,59) (880.4,60) (901.12,61) (936.35,62) (1024.97,63) (1202.96,64) (1228.53,65) (1247.58,66) (1376.42,67) (1482.21,68) (1551.8,69) (1591.21,70) (1700.45,71) (1799.11,72) (2093.54,73) (2198.48,74) (2231.54,75) (2366.72,76) (2556.65,77) (2782.38,78) (2917.06,79) (2938.62,80) (3214.21,81) (3269.34,82) (3420.14,83) (3909.15,84) (4209.18,85) (4260.58,86) (4891.15,87) (10000, 87) };
\addplot[color=colk, mark=\markk] coordinates {(0.01,1) (0.02,2) (0.06,3) (0.07,4) (0.07,5) (0.08,6) (0.12,7) (0.17,8) (0.19,9) (0.31,10) (0.39,11) (0.4,12) (0.42,13) (0.43,14) (0.44,15) (0.46,16) (0.68,17) (1.18,18) (1.42,19) (3.12,20) (4.04,21) (5.18,22) (5.46,23) (5.66,24) (7.38,25) (7.55,26) (9.93,27) (14.08,28) (14.71,29) (15.68,30) (18.92,31) (21.47,32) (21.87,33) (27.91,34) (33.88,35) (34.58,36) (36.51,37) (38.94,38) (40.18,39) (40.23,40) (64.21,41) (69.43,42) (123.84,43) (248.39,44) (291.49,45) (395.63,46) (398.62,47) (398.62,48) (548.01,49) (553.13,50) (605.91,51) (723.87,52) (737.56,53) (806.33,54) (1000.07,55) (1230.11,56) (1302.17,57) (1430.16,58) (1878.4,59) (1922.77,60) (2613.19,61) (2633.72,62) (3200.2,63) (3781.53,64) (3884.67,65)  (10000, 65) };

\legend{ \solverh, \solverg, \solverf, \solverd, \solvere, \solverj, \solveri, \solverc, \solvera, \solverb, \solverk }

\addplot[densely dotted, color=black] coordinates {(5000, -1) (5000, 400)};

          \end{axis}
        \end{tikzpicture}
  
  \caption{\color{black} Solve time comparisons between base \yalsatprob{}, and 10 \yalsatddfw{} settings for \instancesA{} and \instancesB{}, where restarts are enabled}
  \label{fig:comb+sc_res_large}
\end{figure}
\subsection{Evaluation With Restarts}

Many SLS algorithms restart their search with a random assignment after a fixed number of flips.
By default, \yalsat{} also performs restarts.
However, at each restart, \yalsat{} dynamically sets a new restart interval as $r = 100,000x$ for some integer $x \ge 1$, which is initialized to 1, and updated after each restart as follows: if $x$ is power of 2, then $x$ is set to 1, otherwise to $2*x$\color{black}.
The way \yalsat{} initializes its assignment at restart also differs from many SLS algorithms.
On some restarts, \yalsat{} uses the best cached assignment.
For all others, it restarts with a fresh random assignment.
In this way, it attempts to balance exploitation and exploration.

We also evaluated \yalsatddfw{} with \yalsat{}-style restarts.
On a restart, the adjusted clause weights are kept.
The hope is that the adjusted weights help the solver descend the search landscape faster.

We compare \yalsatprob{} against ten experimental settings of \yalsatddfw{} with restarts enabled.
The best solver in this evaluation is \yalsatddfw{} with the setting \linrevs\texttt{-c.1-grdy} on the \instancesA{} benchmark and the setting \linrevs\texttt{-c.1-wrnd} on the \instancesB{} benchmark, which solve 11 and 49 more instances than \yalsatprob{}, respectively.
Figure~\ref{fig:comb+sc_res_large} shows solve counts against solving time, and it confirms that all the \yalsatddfw{} settings solve instances substantially faster than \yalsatprob{}.


\subsection{Solving Hard Instances}

\textbf {Closing wap-07a-40.}
The \wap{} family from the \instancesA{} benchmark contains three open instances: wap-07a-40, wap-03a-40 and wap-4a-40. We attempted to solve these three instances using the parallel version of \yalsatddfw{} with the ten \yalsatddfw{} settings (without restarts) used in Section~\ref{sec-withoutrestart} in the cluster node with 128 cores and 18,000 seconds of timeout.
All of our settings except \texttt{\intwtshort-c.01-\pmbest} (the baseline) solve the wap-07a-40 instance.
The best setting for this experiment was \texttt{\linorigs-c.1-\pmwrand}, which solves wap-07a-40 in just 1168.64 seconds.  
However, we note that lstech\_maple (\texttt{LMpl})~\cite{winner2021}, the winner of the SAT track of the SAT Competition 2021, also solves wap-07a-40, in 2,103.12 seconds, almost twice the time required by our best configuration \texttt{\linorigs-c.1-\pmwrand} for solving this instance. Thus, for solving this open instance, our best setting compares well with the state-of-the-art solver for solving satisfiable instances.

With restarts, the setting \texttt{\linorigs-c.1-\pmwrand}, the best setting for this experiment, were not able to solve any of these three instances.\color{black}

\vspace {.1cm}
\noindent \textbf{New lower bounds for van der Waerden/Green numbers.}
The \textit{van der Waerden theorem}~\cite{vanderweardenthm} is a theorem about the existence of monochromatic arithmetic progressions among a set of numbers.
It states the following: there exists a smallest number $n = W(k; t_1, \dots, t_i, \dots , t_k)$ such that any coloring of the integers $\{1, 2, \dots, n\}$ with $k$ colors contains a progression of length $t_i$ of color $i$ for some $i$.
In recent work, Ben Green showed that these numbers grow much faster than conjectured and that their growth can be observed in experiments~\cite{Green}.
We therefore call the CNF formulas to determine these numbers Green instances. 

Ahmed et al.\ studied 20 van der Waerden numbers $W (2; 3, t)$ for two colors, with the first color having arithmetic progression of length 3 and the second of length $19 \le t \le 39$, and 
conjectured that their values for $t \leq 30$ were optimal, including $W (2;3, 29) = 868$ and $W(2, 3, 30) = 903$~\cite{kullman}.
By using \yalsatddfw{}, we were able to refute these two conjectures by solving the formulas Green-29-868-SAT and Green-30-903-SAT in the \instancesA{} set.
Solving these instances yields two new bounds: $W (2;3, 29) \ge 869$ and $W (2;3, 30) \ge 904$.

To solve these two instances, we ran our various \yalsatddfw{} configurations (without restarts) using \yalsat's parallel mode, along with a number of other local search algorithms from \ubcsat{}, in the same cluster we used to solve wap-07a-40.
Among these solvers, only our solver could solve the two instances.
\texttt{\linorigs-c.1-\pmwrand} solved both Green-29-868-SAT and Green-30-903-SAT, in 942.60 and 6534.56 seconds, respectively.
The settings \texttt{\linrevs-c.1-\pmwrand} and \texttt{\lineqs-c.1-\pmwrand} also solved Green-29-868-SAT in 1374.74 and 1260.16 seconds, respectively, but neither could solve Green-30-903-SAT within a timeout of 18,000 seconds. 
The CDCL solver \texttt{LMpl}, which solves wap-07a-40, could not solve any instances from the Green family within a timeout of 18,000 seconds.

With restarts \texttt{\linorigs-c.1-\pmwrand}, the best setting for this experiment only solves Green-29-868-SAT in 2782.81 seconds within  a timeout of 18,000 seconds. \color{black}
\section{Discussion and Future Work}\label{sec:conclusion}

In this paper, we proposed three modifications to the DLS SAT-solving algorithm \ddfw{}.
We then implemented \ddfw{} on top of the SLS solver \yalsat{} to create the solver \yalsatddfw{}, and we tested this solver on a pair of challenging benchmark sets.
Our experimental results showed that our modifications led to substantial improvement over the baseline \ddfw{} algorithm.
The results show that future users of \yalsatddfw{} should, by default, use the configuration \texttt{\linrevs-c.1-\pmwrand}. 

While each modification led to improved performance, the improvements due to each modification were not equal.
The performance boost due to switching to the weighted-random variable selection method was the weakest, as it resulted in the fewest additional solves.
However, our results indicate that making occasional non-optimal flips may help \ddfw{} explore its search space better.

The performance boost due to adjusting the \cspt{} value was more substantial, supporting our initial findings in Section~\ref{subsec:cspt}.
One metric that could explain the importance of a higher \cspt{} value is a clause's \emph{degree of satisfaction} (DS), which is the fraction of its literals that are satisfied by the current assignment.
We noticed in experiments on the COMB benchmark with \cspt{} = 0.01 that clauses neighboring a falsified clause had an average DS value of 0.33, while clauses without a neighboring falsified clause had an average DS value of 0.54.
If this trend holds for general \yalsatddfw{} runs, then it may be advantageous to take weight from the latter clauses more often, since flipping any literal in a falsified clause will not falsify any of the latter clauses.
A higher \cspt{} value accomplishes this.
However, we did not investigate the relationship between DS and \cspt{} further, and we leave this to future work.
Performance also improved with the switch to a linear weight transfer method.
The best method, \linrevs{}, supports the findings from the workshop paper that \ddfw{} should transfer more weight from clauses with the initial weight.
Future work can examine whether the heavy-clause distinction is valuable;
a weight transfer rule that doesn't explicitly check if a clause is heavy would simplify the \ddfw{} algorithm.

 When restarts are enabled, all ten settings  in \yalsatddfw{} perform better for \instancesA{} than when restarts are disabled. This better performance with restarts comes from solving several \mm{} instances, for which these settings without restarts solve none of them. 
 However, for \instancesB{}, \yalsatddfw{} performs better when restarts are disabled. Since \instancesB{} comprises of substantially larger number of heterogeneous benchmarks than \instancesA{}, these results suggest that the new system performs better when restarts are disabled.
\color{black}

Future work on weight transfer methods can take several other directions.
Different transfer functions can be tested, such as those that are a function of the falsified clause's weight or those based on rational or exponential functions.
Alternate definitions for neighboring clauses are also possible.
For example, in formulas with large neighborhoods, it may be advantageous to consider clauses neighbors if they share $k > 1$ literals, rather than just 1.

\begin{figure}[t]
  \centering
  \begin{tikzpicture}
    \begin{axis}[
      title = {},
      xlabel = {Flips Count},
      ylabel style={align=center},
      ylabel = {Sideways moves count\\per 10,000 flips},
      xmin = 0,
      xmax = 5000000,
      width=0.95\textwidth,
      height=120pt,
      grid = major,
      legend entries = {\solvera, \solverh},]
      \addplot[only marks,mark size=.8pt, color=blue] table {data/sideways_comparison/fw-0.01-grdy.tsv};
      \addplot[only marks,mark size=.8pt, color=red] table {data/sideways_comparison/lwhle-0.1-wrand.tsv};
    \end{axis}
  \end{tikzpicture}
  \caption{Comparison of sideways moves count per 10,000 flips with Search Progression for our baseline (\texttt{fw-c.01-grdy}) and best setting (\texttt{\linrevs{}-c.1-\pmwrand{}})  from \yalsatddfw{} for an \instancesA{} instance sted2\_0x0\_n219-342 }
  \label{fig:sideways}
\end{figure}
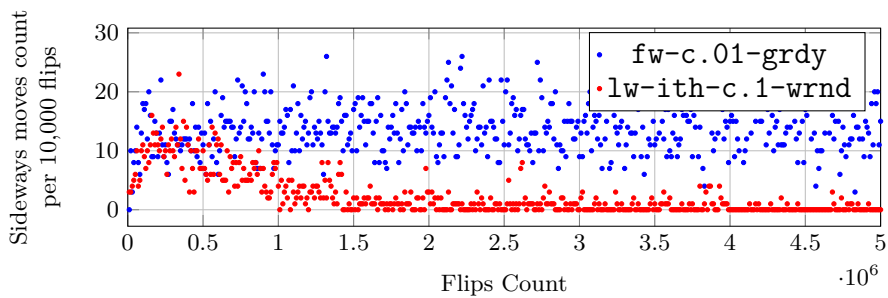
Throughout this paper, we kept the \spt{} parameter set to 0.15.
Yet, when clause weights are floating point numbers, it is rare for our solver to make sideways moves. This evident in Figure~\ref{fig:sideways}, which compares count of sideways moves per 10,000 flips between our baseline setting (\texttt{fw-0.01-grdy}), and best setting (\texttt{\linrevs-c.1-wrand}) for a randomly chosen \instancesB{} instance sted2\_0x0\_n219-342 up to 5 millions flips. With  \texttt{fw-0.01-grdy}, \yalsatddfw{} makes some sideways moves, albeit rarely. However, with floating weight transfer in \texttt{\linrevs-c.1-wrand}, the solver makes almost no sideways moves as search progresses.
We further investigated the effect of sideways moves on solver performance.
We tested the setting \texttt{\linrevs-c.1-\pmwrand} against a version that did not perform sideways moves on the \instancesB{} benchmark.
The version with sideways moves solved 118 instances, while the version without them solved 113.
This suggests that sideways moves may add a slight-but-beneficial amount of random noise to the algorithm.
Future work can more fully investigate the effect of sideways moves on \ddfw{}.
One goal is to eliminate the parameter entirely in order to simplify the algorithm.
Alternatively, the algorithm could be modified to occasionally flip variables that \emph{increase} the falsified weight to help \ddfw{} explore the search space.

Overall, we find that the \ddfw{} algorithm continues to show promise and deserves more research interest.
Our solver closed several hard instances that eluded other state-of-the-art solvers, and the space of potential algorithmic improvements remains rich.

\begin{subappendices}

\section{Experimental Results with a Different Seed}
\label{appendix:A}
For all our previous experiments, we used the default seed (seed of 0) used in the solvers. We have repeated the experiments reported in Section 6.1 and 6.2 for \instancesA{} and \instancesB{} with a seed of 123. Here, we present the results with this changed seed value. 

Figure \ref{fig:comb+sclarge_seed123} and \ref{fig:comb+sc_res_large_seed123} compare the performance of various configurations of \yalsatddfw{} against their baselines for this changed seed, without restarts and with restarts, respectively.  With the changed seeds, when \yalsatddfw{} does not perform restarts (Figure \ref{fig:comb+sclarge_seed123}), all of our configurations performs better than the baseline \texttt{fw-c.01-grdy}, except \texttt{fw-c.01-wrnd}. Similar to the results with seed 0, with the seed of 123, the configurations implementing \linwt{} dominates over the configurations with \intwt{}. When restarts are enabled  (Figure \ref{fig:comb+sc_res_large_seed123}), with seed 123, the overall performance of our configurations are similar to what they are with seed 0 (Figure 4).

\begin{figure}
  \centering
      \begin{tikzpicture}[scale = 1]
          \begin{axis}[mark size=1.5pt, grid=both, grid style={black!10}, legend columns=2, legend style={nodes={scale=0.45]}, at={(0.95,0.32)}}, legend cell align={left}, 
                              ylabel near ticks, ylabel near ticks, ylabel shift={0pt}, width=185pt, height=200pt,xlabel=solving time (s), ylabel=solved COMB instances,xmin=0,xmax=18600,ymin=0,ymax=30]
          
\addplot[color=colh, mark=\markh] coordinates {(1.28,1) (7.49,2) (13.96,3) (43.78,4) (64.27,5) (76.83,6) (77.24,7) (91.16,8) (128.33,9) (142.89,10) (174.49,11) (187.61,12) (261.34,13) (350.44,14) (511.74,15) (574.46,16) (605.34,17) (737.96,18) (976.49,19) (1105.28,20) (1495.93,21) (3205.67,22) (3950.28,23) (5231.82,24) (5685.45,25) (6426.67,26) (9294.62,27) (12743.49,28) (19000, 28) };
\addplot[color=coli, mark=\marki] coordinates { (5.5,1) (9.55,2) (15.02,3) (20.1,4) (21.2,5) (46.03,6) (64.9,7) (87.83,8) (102.06,9) (116.75,10) (165.09,11) (166.47,12) (215.3,13) (225.71,14) (355.52,15) (425.48,16) (677.49,17) (712.73,18) (1021.94,19) (1265.52,20) (1520.6,21) (1625.27,22) (1690.75,23) (4131.88,24) (5938.1,25) (7244.84,26) (7563.6,27) (9007.67,28) (19000, 28)};
\addplot[color=cole, mark=\marke] coordinates { (5.49,1) (9.91,2) (9.98,3) (13.86,4) (60.57,5) (61.31,6) (80.65,7) (105.81,8) (138.47,9) (147.64,10) (154.63,11) (166.71,12) (218.11,13) (229.38,14) (254.66,15) (261.54,16) (346.23,17) (379.94,18) (481.6,19) (513.95,20) (589.38,21) (876.35,22) (1097.77,23) (2777.4,24) (3001.56,25) (9515.45,26) (14482.29,27) (17685.56,28) (19000, 28)};
\addplot[color=colg, mark=\markg] coordinates { (5.39,1) (7.53,2) (18.62,3) (20.55,4) (48.05,5) (57.44,6) (105.11,7) (107.76,8) (128.55,9) (128.72,10) (193.04,11) (206.42,12) (210.17,13) (211.1,14) (340.73,15) (550.51,16) (649.75,17) (697.52,18) (1028.36,19) (1161.7,20) (1435.48,21) (1619.87,22) (2169.89,23) (3175.27,24) (3783.48,25) (7133.47,26) (9528.6,27) (17197.25,28) (19000, 28)};
\addplot[color=colj, mark=\markj] coordinates { (7.53,1) (19.37,2) (22.26,3) (59.33,4) (93.44,5) (93.84,6) (102.55,7) (109.97,8) (136.58,9) (144.87,10) (190.22,11) (191.06,12) (217.97,13) (225.3,14) (234.27,15) (350.41,16) (384.14,17) (412.08,18) (491.85,19) (654.71,20) (746.3,21) (1044.96,22) (1340.28,23) (1810.31,24) (4039.57,25) (6846.67,26) (10144.2,27) (17033.1,28) (19000, 28)};
\addplot[color=colf, mark=\markf] coordinates { (7.49,1) (14.79,2) (61.24,3) (70.08,4) (72.22,5) (74.72,6) (100.76,7) (126.37,8) (136.55,9) (140.19,10) (172.52,11) (346.38,12) (438.45,13) (439.33,14) (609.53,15) (663.03,16) (985.83,17) (1046.98,18) (1301.0,19) (1337.85,20) (1359.54,21) (1538.55,22) (2401.75,23) (9986.25,24) (12109.14,25) (17051.63,26) (19000, 26)};
\addplot[color=colc, mark=\markc] coordinates { (1.74,1) (5.49,2) (10.02,3) (32.29,4) (75.03,5) (87.54,6) (160.65,7) (203.95,8) (221.17,9) (262.87,10) (300.52,11) (487.04,12) (624.3,13) (721.81,14) (761.73,15) (801.31,16) (838.66,17) (2048.29,18) (2089.24,19) (2156.39,20) (2260.35,21) (2482.86,22) (3004.23,23) (3006.18,24) (3298.82,25) (19000, 25)};
\addplot[color=cold, mark=\markd] coordinates { (7.54,1) (14.69,2) (15.01,3) (49.79,4) (53.99,5) (91.65,6) (174.07,7) (203.56,8) (295.39,9) (323.56,10) (348.92,11) (374.37,12) (433.03,13) (633.9,14) (710.61,15) (760.76,16) (781.63,17) (963.01,18) (1033.15,19) (1737.68,20) (1910.31,21) (1990.9,22) (3461.23,23) (15207.3,24) (19000, 24)};
\addplot[color=cola, mark=\marka] coordinates { (5.53,1) (15.73,2) (22.08,3) (58.42,4) (122.43,5) (164.36,6) (596.27,7) (754.0,8) (1081.2,9) (1286.97,10) (2030.14,11) (2036.04,12) (3285.56,13) (3934.65,14) (4864.15,15) (4932.67,16) (5203.23,17) (7476.38,18) (7594.17,19) (9910.66,20) (12241.03,21) (13417.84,22) (15131.97,23) (16379.53,24) (19000, 24)};
\addplot[color=colb, mark=\markb] coordinates { (7.54,1) (21.3,2) (185.79,3) (216.79,4) (234.23,5) (249.36,6) (251.11,7) (359.03,8) (463.85,9) (484.45,10) (566.52,11) (629.58,12) (1035.21,13) (2017.22,14) (2060.6,15) (2898.61,16) (3848.74,17) (4903.77,18) (6587.92,19) (6712.72,20) (7583.67,21) (11477.95,22) (16229.91,23) (17966.81,24) (19000, 24)};

\legend{ \solverh, \solveri, \solvere, \solverg, \solverj,\solverf, \solverc,\solverd, \solvera, \solverb }

\addplot[densely dotted, color=black] coordinates {(18000, -1) (18000, 400)};

          \end{axis}
        \end{tikzpicture}~
  \begin{tikzpicture}[scale = 1]
          \begin{axis}[mark size=1.5pt, grid=both, grid style={black!10}, legend columns=2, legend style={nodes={scale=0.45]}, at={(0.95,0.32)}}, legend cell align={left}, 
                              ylabel near ticks, ylabel near ticks, ylabel shift={-3pt}, width=185pt, height=200pt,xlabel=solving time (s), ylabel=solved SATComp instances,xmin=0,xmax=5200,ymin=0,ymax=120]
          
\addplot[color=colg, mark=\markg] coordinates { (0.01,1) (0.01,2) (0.02,3) (0.02,4) (0.02,5) (0.03,6) (0.09,7) (0.17,8) (0.18,9) (0.2,10) (0.24,11) (0.25,12) (0.27,13) (0.3,14) (0.41,15) (0.42,16) (0.48,17) (0.64,18) (0.71,19) (0.87,20) (0.96,21) (1.0,22) (1.0,23) (1.18,24) (1.3,25) (1.35,26) (1.57,27) (2.12,28) (2.68,29) (4.46,30) (6.51,31) (8.0,32) (8.77,33) (9.76,34) (10.43,35) (12.74,36) (13.73,37) (14.7,38) (15.04,39) (17.53,40) (19.1,41) (19.11,42) (20.59,43) (20.61,44) (22.01,45) (25.4,46) (29.02,47) (34.32,48) (38.33,49) (39.1,50) (42.98,51) (56.48,52) (59.84,53) (60.34,54) (70.63,55) (70.98,56) (76.54,57) (87.85,58) (101.12,59) (115.92,60) (122.5,61) (128.91,62) (134.29,63) (142.4,64) (148.13,65) (181.45,66) (194.68,67) (214.42,68) (233.73,69) (256.41,70) (265.8,71) (272.14,72) (277.79,73) (302.84,74) (335.86,75) (356.34,76) (465.02,77) (495.15,78) (513.15,79) (557.47,80) (559.24,81) (627.71,82) (651.88,83) (654.38,84) (661.88,85) (727.92,86) (770.77,87) (782.86,88) (793.33,89) (900.34,90) (908.57,91) (982.13,92) (1126.72,93) (1154.23,94) (1164.95,95) (1277.37,96) (1297.7,97) (1377.38,98) (1548.88,99) (1671.41,100) (1730.35,101) (2044.52,102) (2105.35,103) (2266.53,104) (2308.25,105) (2481.23,106) (3156.94,107) (3309.56,108) (3316.08,109) (3478.57,110) (3511.03,111) (3557.97,112) (3789.11,113) (3855.9,114) (4038.85,115) (4478.17,116) (4667.17,117) (4908.93,118) (10000, 118) };
\addplot[color=colh, mark=\markh] coordinates { (0.01,1) (0.01,2) (0.01,3) (0.07,4) (0.08,5) (0.09,6) (0.13,7) (0.17,8) (0.18,9) (0.18,10) (0.19,11) (0.24,12) (0.25,13) (0.27,14) (0.27,15) (0.31,16) (0.4,17) (0.44,18) (0.47,19) (0.55,20) (0.62,21) (0.69,22) (1.27,23) (1.34,24) (2.34,25) (2.54,26) (2.86,27) (3.04,28) (3.72,29) (3.73,30) (3.97,31) (4.06,32) (4.84,33) (5.04,34) (5.37,35) (8.3,36) (11.05,37) (11.96,38) (12.86,39) (12.95,40) (15.34,41) (16.04,42) (16.78,43) (17.35,44) (18.46,45) (20.38,46) (21.46,47) (22.81,48) (24.76,49) (32.25,50) (50.73,51) (51.55,52) (55.78,53) (72.63,54) (79.62,55) (87.08,56) (111.22,57) (112.08,58) (121.52,59) (125.54,60) (126.64,61) (127.21,62) (138.81,63) (174.32,64) (195.11,65) (258.89,66) (333.37,67) (385.12,68) (401.97,69) (487.79,70) (538.39,71) (610.37,72) (612.75,73) (688.88,74) (699.05,75) (736.21,76) (797.01,77) (801.76,78) (816.43,79) (854.82,80) (856.34,81) (979.96,82) (982.67,83) (990.52,84) (1097.41,85) (1103.6,86) (1176.67,87) (1281.88,88) (1335.76,89) (1563.13,90) (1601.87,91) (1636.16,92) (1641.9,93) (1648.5,94) (1714.95,95) (2020.01,96) (2230.58,97) (2266.85,98) (2376.1,99) (2392.93,100) (2476.83,101) (2692.29,102) (2727.84,103) (2891.13,104) (3060.19,105) (3115.03,106) (3170.91,107) (3196.73,108) (3286.39,109) (3646.65,110) (4080.45,111) (4106.22,112) (4315.92,113) (4437.27,114) (4567.93,115) (4736.7,116) (4956.77,117) (10000, 117)};
\addplot[color=cole, mark=\marke] coordinates { (0.01,1) (0.01,2) (0.01,3) (0.08,4) (0.1,5) (0.1,6) (0.12,7) (0.22,8) (0.23,9) (0.3,10) (0.31,11) (0.34,12) (0.34,13) (0.38,14) (0.43,15) (0.45,16) (0.48,17) (0.54,18) (0.99,19) (1.1,20) (1.24,21) (1.28,22) (1.4,23) (1.91,24) (3.08,25) (3.26,26) (4.81,27) (5.03,28) (5.2,29) (5.29,30) (6.18,31) (7.56,32) (7.6,33) (8.85,34) (9.2,35) (10.79,36) (11.02,37) (11.3,38) (12.65,39) (13.24,40) (14.2,41) (16.89,42) (17.95,43) (20.51,44) (23.2,45) (23.43,46) (29.86,47) (30.02,48) (34.61,49) (42.03,50) (44.52,51) (55.77,52) (60.8,53) (63.37,54) (68.74,55) (71.8,56) (76.78,57) (81.04,58) (86.67,59) (141.35,60) (148.15,61) (157.81,62) (163.28,63) (170.02,64) (173.77,65) (260.68,66) (291.12,67) (297.79,68) (313.43,69) (325.15,70) (338.14,71) (392.42,72) (481.0,73) (486.56,74) (493.02,75) (611.44,76) (624.52,77) (686.54,78) (690.26,79) (940.99,80) (1037.07,81) (1135.59,82) (1249.73,83) (1515.19,84) (1515.24,85) (1536.69,86) (1656.31,87) (1773.05,88) (1779.62,89) (1801.38,90) (1878.66,91) (1963.14,92) (2063.78,93) (2124.81,94) (2151.0,95) (2356.68,96) (2374.05,97) (2414.64,98) (2678.12,99) (2685.87,100) (2875.01,101) (3134.42,102) (3153.24,103) (3344.56,104) (3382.14,105) (3703.7,106) (4106.27,107) (4312.12,108) (4413.81,109) (4561.74,110) (4700.67,111) (4722.09,112) (10000, 112)};
\addplot[color=colf, mark=\markf] coordinates { (0.0,1) (0.01,2) (0.01,3) (0.06,4) (0.07,5) (0.08,6) (0.08,7) (0.16,8) (0.16,9) (0.21,10) (0.27,11) (0.27,12) (0.42,13) (0.45,14) (0.51,15) (0.66,16) (0.9,17) (0.92,18) (0.93,19) (1.51,20) (1.82,21) (1.92,22) (2.04,23) (2.18,24) (2.78,25) (3.11,26) (4.79,27) (6.57,28) (7.58,29) (7.75,30) (8.79,31) (9.19,32) (9.54,33) (10.5,34) (12.63,35) (13.03,36) (14.84,37) (18.19,38) (19.32,39) (22.76,40) (26.13,41) (27.08,42) (28.93,43) (42.62,44) (65.17,45) (69.14,46) (70.51,47) (79.88,48) (81.24,49) (99.35,50) (104.53,51) (122.76,52) (135.65,53) (136.85,54) (149.09,55) (151.55,56) (163.64,57) (166.57,58) (173.62,59) (183.67,60) (185.19,61) (197.39,62) (207.19,63) (219.45,64) (233.28,65) (238.45,66) (238.55,67) (255.43,68) (283.68,69) (354.06,70) (427.22,71) (445.07,72) (448.89,73) (520.04,74) (588.41,75) (734.97,76) (836.14,77) (943.59,78) (946.76,79) (1003.44,80) (1088.85,81) (1111.45,82) (1164.74,83) (1237.92,84) (1238.33,85) (1245.76,86) (1313.23,87) (1371.84,88) (1935.94,89) (2034.49,90) (2133.23,91) (2174.96,92) (2312.85,93) (2368.64,94) (2384.18,95) (2646.2,96) (2763.16,97) (2834.86,98) (2842.34,99) (3497.44,100) (3522.07,101) (3583.55,102) (3962.83,103) (4110.12,104) (4235.21,105) (4264.06,106) (4657.81,107) (4793.06,108) (4886.46,109) (10000, 109)};
\addplot[color=coli, mark=\marki] coordinates { (0.0,1) (0.0,2) (0.01,3) (0.01,4) (0.02,5) (0.02,6) (0.07,7) (0.08,8) (0.1,9) (0.11,10) (0.14,11) (0.28,12) (0.28,13) (0.31,14) (0.37,15) (0.4,16) (0.48,17) (0.93,18) (1.28,19) (1.32,20) (1.44,21) (1.79,22) (2.14,23) (3.88,24) (4.01,25) (4.16,26) (4.18,27) (4.35,28) (5.13,29) (5.6,30) (8.12,31) (8.53,32) (8.77,33) (9.02,34) (12.07,35) (13.01,36) (13.81,37) (15.49,38) (16.87,39) (22.25,40) (25.54,41) (37.33,42) (38.02,43) (39.09,44) (39.7,45) (41.97,46) (46.67,47) (47.93,48) (66.39,49) (72.98,50) (73.43,51) (77.67,52) (89.94,53) (105.13,54) (111.22,55) (121.57,56) (140.76,57) (202.05,58) (291.29,59) (302.78,60) (311.61,61) (354.62,62) (360.26,63) (368.85,64) (505.36,65) (516.2,66) (542.32,67) (560.36,68) (672.05,69) (675.22,70) (711.5,71) (736.48,72) (1018.1,73) (1215.07,74) (1286.94,75) (1565.44,76) (1604.11,77) (1636.35,78) (1665.82,79) (1684.45,80) (2065.81,81) (2078.03,82) (2105.37,83) (2126.25,84) (2172.7,85) (2258.36,86) (2258.69,87) (2303.18,88) (2734.24,89) (2892.82,90) (2933.69,91) (3157.63,92) (3181.84,93) (3212.76,94) (3397.59,95) (3549.26,96) (3579.16,97) (3636.27,98) (3641.76,99) (3792.52,100) (3816.53,101) (3828.59,102) (3857.07,103) (4038.68,104) (4826.58,105) (4979.17,106) (10000, 106)};
\addplot[color=colj, mark=\markj] coordinates { (0.0,1) (0.02,2) (0.02,3) (0.02,4) (0.05,5) (0.05,6) (0.06,7) (0.08,8) (0.08,9) (0.1,10) (0.21,11) (0.21,12) (0.21,13) (0.4,14) (0.41,15) (0.41,16) (0.51,17) (0.52,18) (1.44,19) (1.52,20) (1.59,21) (1.78,22) (1.96,23) (2.85,24) (3.39,25) (5.02,26) (5.47,27) (6.44,28) (6.55,29) (6.76,30) (10.95,31) (11.93,32) (13.89,33) (14.24,34) (14.6,35) (17.33,36) (17.83,37) (18.26,38) (18.27,39) (20.51,40) (22.7,41) (28.67,42) (39.98,43) (42.95,44) (43.85,45) (46.34,46) (54.77,47) (59.63,48) (61.99,49) (62.44,50) (63.13,51) (71.78,52) (82.7,53) (93.67,54) (93.73,55) (147.26,56) (174.52,57) (174.64,58) (264.49,59) (375.99,60) (379.28,61) (398.06,62) (449.69,63) (484.68,64) (529.72,65) (541.11,66) (649.07,67) (668.45,68) (717.79,69) (770.72,70) (827.67,71) (849.97,72) (853.81,73) (881.62,74) (953.94,75) (969.59,76) (1344.44,77) (1398.26,78) (1406.82,79) (1624.09,80) (1657.99,81) (1723.04,82) (1737.46,83) (1832.18,84) (1842.4,85) (2092.55,86) (2129.55,87) (2333.0,88) (2349.88,89) (2360.15,90) (2447.58,91) (2614.08,92) (3247.49,93) (3460.33,94) (3515.37,95) (3999.25,96) (4003.56,97) (4593.53,98) (4751.98,99) (10000, 99)};
\addplot[color=cold, mark=\markd] coordinates { (0.0,1) (0.0,2) (0.01,3) (0.01,4) (0.02,5) (0.02,6) (0.05,7) (0.12,8) (0.15,9) (0.25,10) (0.28,11) (0.29,12) (0.36,13) (0.42,14) (0.5,15) (0.56,16) (0.71,17) (1.07,18) (1.1,19) (1.46,20) (1.59,21) (2.26,22) (2.28,23) (2.35,24) (2.35,25) (2.47,26) (2.82,27) (3.08,28) (4.33,29) (4.76,30) (5.93,31) (7.56,32) (10.66,33) (13.69,34) (18.64,35) (19.5,36) (22.79,37) (23.17,38) (24.63,39) (29.89,40) (31.25,41) (35.64,42) (37.12,43) (39.33,44) (44.11,45) (57.28,46) (62.02,47) (64.59,48) (69.74,49) (89.61,50) (97.42,51) (104.13,52) (104.46,53) (107.3,54) (113.38,55) (113.72,56) (149.2,57) (164.07,58) (178.14,59) (223.52,60) (226.98,61) (236.82,62) (289.77,63) (354.18,64) (372.71,65) (458.28,66) (766.78,67) (849.47,68) (1027.46,69) (1032.44,70) (1056.02,71) (1226.71,72) (1238.05,73) (1377.34,74) (1416.98,75) (1578.49,76) (1663.94,77) (1747.66,78) (2090.16,79) (2249.68,80) (2280.61,81) (2323.19,82) (2715.25,83) (3074.5,84) (3604.54,85) (3826.67,86) (3999.35,87) (4132.64,88) (4183.27,89) (4281.38,90) (4313.52,91) (4578.19,92) (4727.69,93) (4912.95,94) (4961.03,95) (10000, 95)};
\addplot[color=colc, mark=\markc] coordinates { (0.0,1) (0.01,2) (0.01,3) (0.02,4) (0.07,5) (0.07,6) (0.1,7) (0.18,8) (0.21,9) (0.22,10) (0.24,11) (0.41,12) (0.45,13) (0.48,14) (0.53,15) (0.69,16) (0.75,17) (0.78,18) (1.02,19) (1.11,20) (1.42,21) (1.48,22) (2.98,23) (5.14,24) (5.3,25) (6.31,26) (6.87,27) (9.07,28) (9.3,29) (9.83,30) (9.94,31) (10.26,32) (11.73,33) (13.2,34) (15.02,35) (15.09,36) (15.62,37) (19.26,38) (19.36,39) (21.04,40) (22.17,41) (25.84,42) (34.7,43) (35.67,44) (37.03,45) (37.83,46) (46.07,47) (75.22,48) (96.37,49) (106.08,50) (126.97,51) (128.7,52) (133.51,53) (141.51,54) (168.25,55) (171.96,56) (196.77,57) (200.7,58) (250.54,59) (278.52,60) (284.42,61) (320.32,62) (387.01,63) (410.65,64) (466.43,65) (494.96,66) (637.73,67) (832.73,68) (988.1,69) (990.76,70) (1232.53,71) (1379.84,72) (1434.54,73) (1483.89,74) (1828.31,75) (1944.77,76) (1961.69,77) (1974.51,78) (2071.35,79) (2087.16,80) (2095.33,81) (2108.55,82) (2160.24,83) (2241.08,84) (2536.15,85) (2537.6,86) (2990.83,87) (3033.93,88) (3587.15,89) (3710.92,90) (3717.95,91) (3739.76,92) (4197.62,93) (4998.59,94) (10000, 94)};
\addplot[color=cola, mark=\marka] coordinates { (0.0,1) (0.0,2) (0.01,3) (0.02,4) (0.04,5) (0.05,6) (0.06,7) (0.07,8) (0.15,9) (0.15,10) (0.22,11) (0.41,12) (0.42,13) (0.57,14) (0.92,15) (1.15,16) (1.24,17) (2.1,18) (2.16,19) (2.66,20) (2.69,21) (2.7,22) (3.04,23) (3.47,24) (4.57,25) (4.8,26) (6.43,27) (6.45,28) (7.85,29) (9.26,30) (10.72,31) (11.55,32) (13.73,33) (13.79,34) (14.37,35) (15.45,36) (18.27,37) (21.87,38) (25.42,39) (25.63,40) (37.14,41) (45.72,42) (49.43,43) (50.04,44) (50.81,45) (75.21,46) (108.33,47) (114.24,48) (143.75,49) (195.35,50) (222.15,51) (242.02,52) (298.34,53) (371.39,54) (442.29,55) (461.24,56) (465.08,57) (536.66,58) (574.59,59) (602.98,60) (693.55,61) (889.5,62) (938.08,63) (1050.24,64) (1062.72,65) (1110.25,66) (1150.02,67) (1561.81,68) (1791.7,69) (2035.13,70) (2052.13,71) (2070.56,72) (2216.06,73) (2308.53,74) (2632.73,75) (2941.13,76) (3114.09,77) (3130.09,78) (3210.63,79) (3264.47,80) (3312.65,81) (3337.76,82) (3618.71,83) (4863.27,84) (4904.18,85) (4991.24,86) (10000, 86)};
\addplot[color=colb, mark=\markb] coordinates { (0.03,1) (0.03,2) (0.03,3) (0.06,4) (0.06,5) (0.08,6) (0.09,7) (0.1,8) (0.14,9) (0.17,10) (0.17,11) (0.43,12) (0.61,13) (0.76,14) (0.92,15) (1.2,16) (1.38,17) (1.4,18) (1.58,19) (1.6,20) (2.41,21) (2.44,22) (2.92,23) (3.23,24) (3.38,25) (4.82,26) (10.7,27) (10.77,28) (10.79,29) (17.77,30) (20.73,31) (20.92,32) (21.4,33) (25.54,34) (25.78,35) (34.89,36) (37.91,37) (45.23,38) (48.72,39) (49.91,40) (50.55,41) (60.38,42) (66.95,43) (68.57,44) (93.3,45) (100.0,46) (111.92,47) (122.26,48) (128.07,49) (153.95,50) (159.63,51) (172.8,52) (188.47,53) (209.08,54) (213.52,55) (218.74,56) (246.66,57) (275.64,58) (286.78,59) (310.33,60) (375.87,61) (460.12,62) (463.46,63) (507.98,64) (559.78,65) (573.55,66) (627.76,67) (826.71,68) (864.29,69) (953.84,70) (1106.12,71) (1136.91,72) (1148.03,73) (1525.49,74) (1649.49,75) (2090.98,76) (2278.6,77) (2738.93,78) (3436.2,79) (3752.52,80) (3783.64,81) (4312.22,82) (4540.16,83) (4849.27,84) (10000, 84)};

\legend{ \solverg, \solverh, \solvere, \solverf, \solveri, \solverj, \solverd, \solverc,\solvera, \solverb
 }

\addplot[densely dotted, color=black] coordinates {(5000, -1) (5000, 400)};

          \end{axis}
        \end{tikzpicture}
  
  \caption{Performance profiles of \yalsatddfw{} (fw-c.01-grdy) and nine modifications for \instancesA{} (left) and \instancesB{} (right) with seed 123}
  \label{fig:comb+sclarge_seed123}
\end{figure}
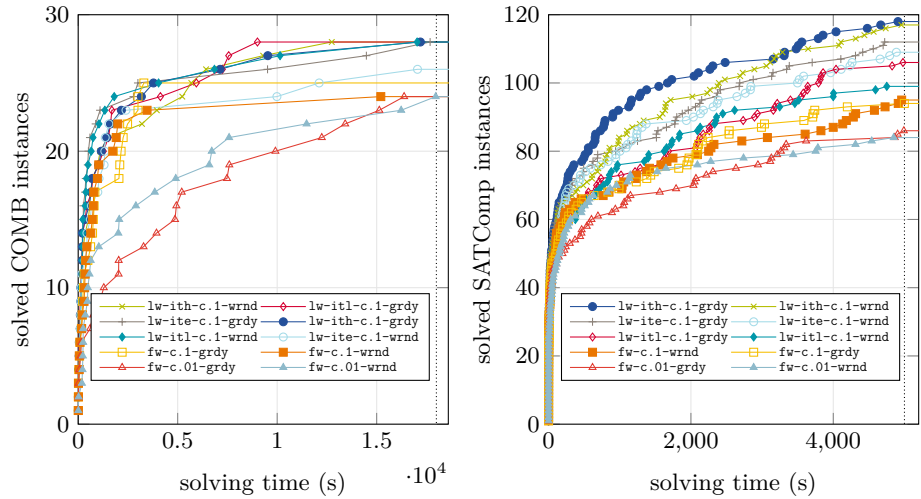

\begin{figure} [h]
  \centering

    \begin{tikzpicture}[scale = 1]
          \begin{axis}[mark size=1.5pt, grid=both, grid style={black!10}, legend columns=2, legend style={nodes={scale=0.45]}, at={(0.95,0.32)}}, legend cell align={left}, 
                              ylabel near ticks, ylabel near ticks, ylabel shift={0pt}, width=185pt, height=200pt,xlabel=solving time (s), ylabel=solved COMP instances,xmin=0,xmax=18600,ymin=0,ymax=43]
          
\addplot[color=colg, mark=\markg] coordinates {  (4.01,1) (5.46,2) (9.89,3) (11.41,4) (13.22,5) (43.23,6) (57.42,7) (98.33,8) (112.42,9) (116.28,10) (192.31,11) (193.4,12) (206.25,13) (216.49,14) (217.72,15) (228.82,16) (308.67,17) (340.16,18) (376.49,19) (440.29,20) (518.39,21) (565.67,22) (686.24,23) (747.43,24) (766.88,25) (898.23,26) (917.07,27) (1017.84,28) (1264.75,29) (2180.67,30) (2233.09,31) (2477.88,32) (3305.78,33) (5223.06,34) (5520.49,35) (9856.99,36) (10930.0,37) (12166.67,38) (15210.91,39) (17813.2,40) (19000,40)};
\addplot[color=colh, mark=\markh] coordinates { (3.76,1) (7.51,2) (9.26,3) (9.56,4) (20.63,5) (43.89,6) (88.55,7) (101.28,8) (108.41,9) (143.3,10) (183.16,11) (183.84,12) (222.61,13) (248.86,14) (321.86,15) (446.68,16) (528.7,17) (564.74,18) (673.36,19) (676.65,20) (749.14,21) (766.86,22) (1064.91,23) (1177.68,24) (1179.65,25) (1456.38,26) (1548.9,27) (2178.6,28) (2197.41,29) (2898.05,30) (3039.15,31) (3938.43,32) (4765.29,33) (7780.76,34) (7966.13,35) (9410.96,36) (12210.59,37) (12641.28,38) (15961.22,39) (19000,39)};
\addplot[color=cola, mark=\marka] coordinates { (5.7,1) (12.51,2) (15.79,3) (69.14,4) (113.58,5) (124.23,6) (162.4,7) (164.2,8) (173.16,9) (346.11,10) (405.82,11) (415.64,12) (429.25,13) (498.42,14) (626.2,15) (694.52,16) (699.32,17) (781.64,18) (973.78,19) (1601.61,20) (1616.87,21) (1672.05,22) (1741.48,23) (1994.96,24) (2426.48,25) (3223.88,26) (3580.18,27) (4264.53,28) (4386.17,29) (4588.83,30) (4857.98,31) (5364.5,32) (6308.31,33) (8442.6,34) (9936.89,35) (10299.53,36) (14968.84,37) (19000,37)};
\addplot[color=cold, mark=\markd] coordinates {  (7.62,1) (15.12,2) (44.0,3) (53.66,4) (66.99,5) (94.21,6) (94.83,7) (111.43,8) (311.99,9) (338.26,10) (409.85,11) (425.96,12) (485.26,13) (575.64,14) (600.8,15) (624.93,16) (657.78,17) (717.46,18) (832.28,19) (932.39,20) (933.95,21) (1053.36,22) (1200.05,23) (1580.02,24) (1868.77,25) (1907.85,26) (1995.8,27) (3354.6,28) (3639.48,29) (5111.67,30) (9130.51,31) (10191.34,32) (12363.5,33) (12448.36,34) (13359.84,35) (13650.17,36) (19000,36)};
\addplot[color=colb, mark=\markb] coordinates { (7.64,1) (25.18,2) (48.1,3) (78.35,4) (90.9,5) (202.35,6) (316.02,7) (357.36,8) (359.79,9) (457.33,10) (568.92,11) (779.82,12) (879.0,13) (1033.61,14) (1368.31,15) (1484.43,16) (1627.44,17) (2012.38,18) (2120.98,19) (2213.1,20) (2330.42,21) (2941.63,22) (3119.55,23) (3260.72,24) (3744.79,25) (3827.38,26) (3955.96,27) (4614.83,28) (4822.32,29) (5353.69,30) (5741.08,31) (6882.92,32) (8198.02,33) (9180.26,34) (9504.45,35) (16350.43,36) (19000,36) };
\addplot[color=cole, mark=\marke] coordinates { (5.5,1) (9.85,2) (13.85,3) (59.91,4) (66.07,5) (80.43,6) (97.6,7) (104.8,8) (105.8,9) (121.0,10) (123.21,11) (142.5,12) (215.49,13) (355.54,14) (403.61,15) (539.14,16) (584.31,17) (643.41,18) (675.8,19) (692.42,20) (705.5,21) (1097.49,22) (1224.0,23) 
(1346.05,24) (1414.71,25) (1838.33,26) (2010.99,27) (2132.55,28) (2236.31,29) (2697.52,30) (7566.21,31) (9039.46,32) (9943.12,33) (11722.42,34) (15550.31,35) (17207.93,36) (19000,36)
};
\addplot[color=colc, mark=\markc] coordinates { (1.81,1) (5.58,2) (7.64,3) (19.24,4) (20.02,5) (60.24,6) (60.61,7) (84.67,8) (160.03,9) (221.12,10) (349.45,11) (460.23,12) (488.86,13) (527.31,14) (626.76,15) (735.09,16) (897.16,17) (1035.96,18) (1086.98,19) (1351.32,20) (1651.13,21) (2047.47,22) (2528.15,23) (2613.23,24) (3049.92,25) (3346.2,26) (3683.67,27) (4183.83,28) (5328.67,29) (9130.39,30) (9663.64,31) (11915.47,32) (14593.14,33) (15509.96,34) (17146.12,35) (19000,35)};
\addplot[color=colf, mark=\markf] coordinates { (7.49,1) (9.88,2) (76.17,3) (81.16,4) (111.09,5) (136.82,6) (164.7,7) (173.83,8) (182.26,9) (193.52,10) (226.85,11) (235.55,12) (246.88,13) (285.37,14) (314.25,15) (345.68,16) (411.23,17) (450.79,18) (451.87,19) (463.37,20) (473.3,21) (495.64,22) (517.9,23) (551.66,24) (744.57,25) (837.66,26) (1159.08,27) (2005.4,28) (5920.72,29) (7404.65,30) (8185.03,31) (12511.64,32) (14151.71,33) (14539.94,34)};
\addplot[color=colj, mark=\markj] coordinates { (7.62,1) (17.17,2) (22.14,3) (26.97,4) (86.67,5) (145.61,6) (155.8,7) (184.35,8) (307.62,9) (311.51,10) (413.32,11) (423.19,12) (458.79,13) (476.07,14) (789.79,15) (881.17,16) (914.44,17) (1207.0,18) (1223.8,19) (1288.82,20) (1366.09,21) (1417.72,22) (1769.73,23) (2177.0,24) (2939.65,25) (3157.17,26) (3192.18,27) (3455.42,28) (3665.29,29) (3681.74,30) (4955.93,31) (5116.41,32) (19000,30)};
\addplot[color=colk, mark=\markk] coordinates {  (0.83,1) (0.99,2) (9.85,3) (22.75,4) (33.67,5) (172.58,6) (191.4,7) (206.31,8) (281.91,9) (352.71,10) (374.5,11) (381.27,12) (510.69,13) (534.42,14) (669.45,15) (690.99,16) (773.37,17) (893.69,18) (1343.0,19) (1411.28,20) (1528.36,21) (2078.84,22) (2599.68,23) (2724.23,24) (2759.67,25) (3585.24,26) (4350.92,27) (5817.25,28) (6285.8,29) (7430.55,30) (19000,30)};

\addplot[color=coli, mark=\marki] coordinates { (5.49,1) (9.41,2) (14.98,3) (20.0,4) (22.58,5) (45.91,6) (64.74,7) (87.74,8) (106.18,9) (114.17,10) (165.18,11) (166.61,12) (206.78,13) (226.54,14) (356.91,15) (425.79,16) (678.53,17) (711.42,18) (1016.8,19) (1267.46,20) (1514.92,21) (1624.27,22) (1691.17,23) (4155.67,24) (5985.63,25) (7193.09,26) (7527.71,27) (9022.57,28) (19000,28)};

\legend{ \solverg, \solverh,  \solvera, \solverd, \solverb, \solvere, \solverc, \solverf, \solverj, \solverk, \solveri }

\addplot[densely dotted, color=black] coordinates {(18000, -1) (18000, 400)};

          \end{axis}
        \end{tikzpicture}~
  \begin{tikzpicture}[scale = 1]
          \begin{axis}[mark size=1.5pt, grid=both, grid style={black!10}, legend columns=2, legend style={nodes={scale=0.45]}, at={(0.95,0.32)}}, legend cell align={left}, 
                              ylabel near ticks, ylabel near ticks, ylabel shift={-3pt}, width=185pt, height=200pt,xlabel=solving time (s), ylabel=solved SATComp instances,xmin=0,xmax=5200,ymin=0,ymax=125]
          
\addplot[color=cole, mark=\marke] coordinates { (0.0,1) (0.01,2) (0.01,3) (0.08,4) (0.09,5) (0.1,6) (0.1,7) (0.12,8) (0.21,9) (0.23,10) (0.34,11) (0.34,12) (0.37,13) (0.43,14) (0.48,15) (0.5,16) (0.99,17) (1.14,18) (1.15,19) (1.22,20) (3.24,21) (3.45,22) (3.5,23) (4.07,24) (4.73,25) (4.81,26) (5.18,27) (5.3,28) (5.62,29) (10.72,30) (11.08,31) (11.88,32) (13.61,33) (14.38,34) (14.58,35) (14.92,36) (15.97,37) (18.12,38) (19.58,39) (20.04,40) (21.05,41) (23.67,42) (28.93,43) (39.38,44) (41.27,45) (44.26,46) (49.27,47) (52.37,48) (54.44,49) (55.51,50) (66.76,51) (67.94,52) (69.37,53) (69.73,54) (77.99,55) (81.2,56) (82.82,57) (98.35,58) (116.04,59) (121.11,60) (131.22,61) (138.6,62) (139.34,63) (140.88,64) (151.75,65) (163.7,66) (185.77,67) (188.59,68) (219.39,69) (243.47,70) (366.53,71) (389.24,72) (485.9,73) (537.22,74) (576.29,75) (769.01,76) (851.98,77) (874.4,78) (998.93,79) (1176.32,80) (1367.22,81) (1407.41,82) (1608.32,83) (1629.79,84) (1830.75,85) (1832.62,86) (1854.84,87) (1975.67,88) (2138.64,89) (2169.63,90) (2539.3,91) (2731.97,92) (2738.31,93) (2857.91,94) (3189.62,95) (3395.48,96) (3431.13,97) (3520.78,98) (3597.82,99) (3796.97,100) (3909.28,101) (4120.66,102) (4620.37,103) (4670.05,104) (4706.43,105)  (10000, 105)};
\addplot[color=colg, mark=\markg] coordinates { (0.01,1) (0.02,2) (0.02,3) (0.02,4) (0.02,5) (0.03,6) (0.05,7) (0.09,8) (0.18,9) (0.2,10) (0.2,11) (0.24,12) (0.24,13) (0.27,14) (0.29,15) (0.3,16) (0.36,17) (0.42,18) (0.48,19) (0.68,20) (0.84,21) (0.98,22) (1.01,23) (1.16,24) (1.17,25) (1.39,26) (1.9,27) (2.02,28) (3.52,29) (3.99,30) (4.23,31) (11.17,32) (12.02,33) (13.58,34) (13.62,35) (14.38,36) (14.47,37) (15.56,38) (16.23,39) (16.44,40) (17.64,41) (27.25,42) (30.69,43) (31.69,44) (42.87,45) (43.33,46) (44.27,47) (44.8,48) (46.61,49) (58.32,50) (75.31,51) (83.82,52) (87.93,53) (93.88,54) (102.35,55) (133.47,56) (141.14,57) (169.42,58) (184.75,59) (209.99,60) (214.97,61) (236.5,62) (264.0,63) (266.13,64) (274.07,65) (284.41,66) (289.79,67) (302.38,68) (318.47,69) (430.32,70) (431.71,71) (439.76,72) (441.75,73) (442.36,74) (508.71,75) (518.23,76) (519.83,77) (629.98,78) (671.04,79) (867.81,80) (924.4,81) (1090.69,82) (1376.84,83) (1495.89,84) (1498.74,85) (1550.57,86) (1815.61,87) (2074.76,88) (2112.75,89) (2196.47,90) (2233.59,91) (2295.75,92) (2670.01,93) (2858.54,94) (2911.4,95) (3243.61,96) (3323.52,97) (3385.05,98) (3435.33,99) (3462.13,100) (3528.22,101) (4050.97,102) (4407.37,103) (4491.7,104) (4944.74,105)  (10000, 105)};
\addplot[color=colh, mark=\markh] coordinates { (0.0,1) (0.01,2) (0.01,3) (0.07,4) (0.08,5) (0.09,6) (0.1,7) (0.12,8) (0.12,9) (0.13,10) (0.16,11) (0.18,12) (0.25,13) (0.26,14) (0.27,15) (0.31,16) (0.43,17) (0.44,18) (0.46,19) (0.62,20) (0.77,21) (0.81,22) (1.47,23) (2.03,24) (2.53,25) (2.68,26) (3.75,27) (3.76,28) (3.98,29) (4.83,30) (5.26,31) (5.67,32) (8.06,33) (9.47,34) (10.23,35) (12.21,36) (13.4,37) (15.41,38) (16.08,39) (17.78,40) (20.51,41) (20.61,42) (25.79,43) (33.11,44) (36.27,45) (37.36,46) (46.21,47) (48.71,48) (55.86,49) (61.65,50) (65.29,51) (69.4,52) (78.88,53) (80.97,54) (138.83,55) (139.5,56) (142.97,57) (143.81,58) (145.73,59) (149.4,60) (151.6,61) (163.48,62) (173.3,63) (185.87,64) (193.74,65) (201.96,66) (203.93,67) (235.07,68) (280.98,69) (290.25,70) (377.56,71) (381.31,72) (396.91,73) (425.02,74) (470.55,75) (484.13,76) (501.13,77) (623.71,78) (657.36,79) (743.4,80) (905.85,81) (944.64,82) (983.57,83) (1127.45,84) (1151.1,85) (1333.49,86) (1389.63,87) (1612.11,88) (1616.56,89) (1997.68,90) (2337.55,91) (2397.16,92) (2619.45,93) (2748.18,94) (2801.0,95) (3110.49,96) (3646.28,97) (3693.1,98) (3797.41,99) (3804.21,100) (3945.63,101) (4051.76,102) (4169.4,103) (4697.84,104)  (10000, 104)};
\addplot[color=colf, mark=\markf] coordinates { (0.0,1) (0.01,2) (0.02,3) (0.06,4) (0.07,5) (0.08,6) (0.08,7) (0.08,8) (0.1,9) (0.16,10) (0.16,11) (0.21,12) (0.27,13) (0.42,14) (0.43,15) (0.44,16) (0.56,17) (0.9,18) (0.98,19) (1.57,20) (2.04,21) (3.38,22) (3.58,23) (4.98,24) (6.56,25) (9.54,26) (9.55,27) (9.68,28) (9.88,29) (10.86,30) (12.53,31) (16.51,32) (16.7,33) (17.86,34) (21.05,35) (21.56,36) (21.66,37) (21.79,38) (23.04,39) (29.35,40) (30.66,41) (31.45,42) (34.15,43) (37.59,44) (40.27,45) (41.28,46) (45.9,47) (49.19,48) (55.71,49) (57.56,50) (59.26,51) (70.16,52) (80.91,53) (84.12,54) (85.22,55) (88.07,56) (108.3,57) (127.07,58) (133.57,59) (135.55,60) (173.05,61) (197.29,62) (198.17,63) (245.77,64) (273.81,65) (284.25,66) (296.92,67) (344.55,68) (372.48,69) (408.56,70) (438.98,71) (511.3,72) (541.65,73) (546.21,74) (710.25,75) (827.32,76) (956.9,77) (1057.22,78) (1203.9,79) (1549.26,80) (1572.74,81) (1677.67,82) (1740.95,83) (1835.93,84) (1893.53,85) (2157.87,86) (2310.4,87) (2323.18,88) (2423.75,89) (2856.41,90) (2858.43,91) (3070.3,92) (3250.15,93) (3334.53,94) (3476.57,95) (3478.84,96) (3731.23,97) (3738.72,98) (3933.82,99) (4462.67,100) (4744.64,101) (4755.61,102)  (10000, 102)};
\addplot[color=colc, mark=\markc] coordinates { (0.0,1) (0.01,2) (0.01,3) (0.01,4) (0.07,5) (0.08,6) (0.1,7) (0.12,8) (0.18,9) (0.21,10) (0.22,11) (0.23,12) (0.36,13) (0.44,14) (0.45,15) (0.5,16) (0.69,17) (0.75,18) (1.09,19) (1.23,20) (1.48,21) (1.84,22) (1.89,23) (2.96,24) (2.98,25) (4.02,26) (5.36,27) (8.07,28) (8.69,29) (9.14,30) (10.45,31) (13.69,32) (16.1,33) (18.57,34) (22.54,35) (30.51,36) (32.85,37) (37.17,38) (38.4,39) (40.58,40) (48.98,41) (49.34,42) (67.46,43) (69.71,44) (79.71,45) (82.57,46) (86.42,47) (95.66,48) (120.13,49) (123.76,50) (141.36,51) (159.32,52) (172.05,53) (181.73,54) (196.58,55) (225.59,56) (241.35,57) (282.16,58) (336.28,59) (338.73,60) (343.39,61) (365.81,62) (483.3,63) (516.0,64) (584.11,65) (599.0,66) (677.51,67) (678.84,68) (711.43,69) (714.69,70) (718.67,71) (722.51,72) (725.61,73) (727.68,74) (846.76,75) (914.41,76) (940.05,77) (1023.7,78) (1024.27,79) (1027.29,80) (1035.22,81) (1112.42,82) (1392.82,83) (1561.81,84) (1573.01,85) (1622.03,86) (2041.27,87) (2101.48,88) (2573.94,89) (2685.93,90) (3047.77,91) (3208.44,92) (3856.73,93) (4106.61,94) (4152.93,95) (4208.79,96) (4554.14,97) (4650.89,98)  (10000, 98)};
\addplot[color=cold, mark=\markd] coordinates { (0.0,1) (0.01,2) (0.01,3) (0.02,4) (0.02,5) (0.02,6) (0.06,7) (0.08,8) (0.12,9) (0.15,10) (0.19,11) (0.28,12) (0.29,13) (0.3,14) (0.33,15) (0.36,16) (0.45,17) (0.47,18) (0.67,19) (1.1,20) (1.2,21) (1.38,22) (1.59,23) (1.6,24) (1.61,25) (1.76,26) (2.34,27) (2.42,28) (2.46,29) (4.28,30) (6.15,31) (6.23,32) (7.03,33) (8.97,34) (12.55,35) (17.93,36) (22.12,37) (22.99,38) (31.18,39) (32.87,40) (40.04,41) (44.65,42) (56.74,43) (75.55,44) (91.49,45) (94.58,46) (106.96,47) (109.58,48) (110.89,49) (116.72,50) (173.1,51) (183.51,52) (191.9,53) (210.65,54) (218.42,55) (221.86,56) (230.71,57) (304.03,58) (325.18,59) (328.02,60) (380.8,61) (381.37,62) (411.07,63) (412.35,64) (423.43,65) (448.75,66) (477.91,67) (545.2,68) (553.18,69) (559.94,70) (576.04,71) (576.52,72) (595.9,73) (601.65,74) (615.65,75) (701.81,76) (727.49,77) (775.79,78) (961.17,79) (1109.72,80) (1291.16,81) (1603.76,82) (1889.04,83) (1957.02,84) (2097.2,85) (2185.5,86) (2314.98,87) (2415.73,88) (2716.62,89) (2802.79,90) (3532.19,91) (3596.35,92) (3697.13,93) (4032.1,94) (4078.05,95) (4761.21,96)  (10000, 96)};
\addplot[color=coli, mark=\marki] coordinates { (0.0,1) (0.0,2) (0.02,3) (0.02,4) (0.02,5) (0.02,6) (0.07,7) (0.1,8) (0.11,9) (0.13,10) (0.14,11) (0.14,12) (0.24,13) (0.26,14) (0.28,15) (0.37,16) (0.45,17) (0.5,18) (0.82,19) (0.84,20) (1.32,21) (1.4,22) (2.14,23) (2.15,24) (2.29,25) (3.12,26) (3.16,27) (5.13,28) (5.58,29) (6.53,30) (8.96,31) (9.23,32) (9.58,33) (11.37,34) (11.64,35) (13.76,36) (19.09,37) (19.21,38) (19.75,39) (30.29,40) (34.18,41) (36.3,42) (41.41,43) (43.42,44) (43.87,45) (44.09,46) (48.36,47) (51.84,48) (59.89,49) (75.4,50) (104.87,51) (105.98,52) (111.0,53) (116.25,54) (128.63,55) (141.0,56) (145.69,57) (157.69,58) (157.94,59) (167.15,60) (194.14,61) (194.87,62) (210.15,63) (211.41,64) (215.24,65) (232.4,66) (240.89,67) (371.36,68) (439.12,69) (461.88,70) (465.53,71) (548.59,72) (572.95,73) (592.22,74) (660.82,75) (660.85,76) (707.56,77) (709.51,78) (755.65,79) (886.29,80) (1076.99,81) (1142.6,82) (1474.88,83) (1570.34,84) (1603.12,85) (1629.56,86) (1893.11,87) (2075.42,88) (2308.31,89) (2496.7,90) (2630.45,91) (3181.01,92) (3221.53,93) (3725.88,94) (3841.95,95) (4748.66,96)  (10000, 96)};
\addplot[color=colj, mark=\markj] coordinates { (0.0,1) (0.01,2) (0.02,3) (0.02,4) (0.05,5) (0.05,6) (0.06,7) (0.08,8) (0.08,9) (0.1,10) (0.21,11) (0.21,12) (0.21,13) (0.29,14) (0.4,15) (0.41,16) (0.42,17) (0.45,18) (0.51,19) (0.53,20) (1.5,21) (2.0,22) (2.3,23) (3.88,24) (4.62,25) (5.01,26) (6.66,27) (7.23,28) (7.67,29) (8.47,30) (10.32,31) (10.88,32) (14.7,33) (17.22,34) (17.68,35) (17.76,36) (17.81,37) (17.87,38) (17.96,39) (23.83,40) (24.07,41) (34.87,42) (35.34,43) (41.28,44) (44.01,45) (64.27,46) (86.01,47) (102.34,48) (124.87,49) (151.42,50) (166.4,51) (173.98,52) (177.78,53) (192.04,54) (194.99,55) (216.54,56) (229.17,57) (238.74,58) (245.06,59) (256.68,60) (285.56,61) (312.62,62) (329.71,63) (341.5,64) (384.34,65) (398.48,66) (422.06,67) (459.45,68) (463.39,69) (466.35,70) (603.31,71) (638.19,72) (646.74,73) (654.64,74) (686.41,75) (715.32,76) (741.2,77) (754.62,78) (901.6,79) (936.71,80) (993.6,81) (1035.37,82) (1602.53,83) (2297.61,84) (2353.01,85) (3272.32,86) (3429.95,87) (3560.5,88) (3724.83,89) (3865.81,90) (4254.0,91) (4301.76,92) (4828.65,93)  (10000, 93) };
\addplot[color=cola, mark=\marka] coordinates { (0.02,1) (0.03,2) (0.03,3) (0.05,4) (0.06,5) (0.08,6) (0.09,7) (0.1,8) (0.14,9) (0.17,10) (0.17,11) (0.2,12) (0.38,13) (0.44,14) (0.45,15) (0.6,16) (0.76,17) (0.92,18) (1.39,19) (1.57,20) (3.23,21) (3.95,22) (4.25,23) (4.85,24) (6.82,25) (8.5,26) (9.19,27) (13.51,28) (13.93,29) (14.42,30) (15.88,31) (18.32,32) (19.55,33) (21.04,34) (30.06,35) (30.96,36) (33.07,37) (33.63,38) (36.37,39) (53.82,40) (68.23,41) (70.27,42) (72.24,43) (89.76,44) (110.14,45) (113.02,46) (119.38,47) (126.82,48) (129.15,49) (132.25,50) (173.37,51) (173.59,52) (197.29,53) (211.52,54) (223.86,55) (251.16,56) (275.48,57) (341.7,58) (375.87,59) (393.73,60) (465.7,61) (608.58,62) (675.94,63) (745.45,64) (823.85,65) (913.3,66) (1031.72,67) (1100.44,68) (1175.01,69) (1216.6,70) (1379.72,71) (1469.67,72) (1616.16,73) (2127.08,74) (2236.96,75) (2319.06,76) (2389.52,77) (2472.43,78) (2584.32,79) (2593.68,80) (2728.93,81) (2764.74,82) (3043.62,83) (3113.78,84) (3144.0,85) (4341.36,86) (4516.82,87) (4910.88,88) (4917.12,89)  (10000, 89)};
\addplot[color=colk, mark=\markk] coordinates { (0.0,1) (0.0,2) (0.0,3) (0.01,4) (0.04,5) (0.04,6) (0.05,7) (0.07,8) (0.14,9) (0.15,10) (0.16,11) (0.37,12) (0.42,13) (0.45,14) (0.59,15) (0.71,16) (0.92,17) (0.94,18) (1.1,19) (1.15,20) (1.38,21) (1.96,22) (2.19,23) (2.66,24) (2.71,25) (3.59,26) (5.72,27) (6.41,28) (6.41,29) (8.51,30) (8.71,31) (14.13,32) (14.45,33) (17.39,34) (18.3,35) (20.99,36) (25.62,37) (25.68,38) (36.84,39) (37.75,40) (52.07,41) (53.73,42) (57.76,43) (66.81,44) (66.96,45) (76.13,46) (90.92,47) (103.35,48) (122.6,49) (124.31,50) (141.16,51) (159.57,52) (177.32,53) (217.49,54) (218.78,55) (332.1,56) (398.69,57) (425.72,58) (427.04,59) (736.62,60) (831.29,61) (879.25,62) (1033.14,63) (1035.44,64) (1048.03,65) (1201.49,66) (1242.51,67) (1249.71,68) (1274.22,69) (1464.44,70) (1620.39,71) (1624.62,72) (1959.96,73) (2118.79,74) (2121.09,75) (2505.08,76) (3150.84,77) (3201.51,78) (3206.12,79) (3287.32,80) (4046.79,81) (4405.89,82) (4503.8,83) (4971.36,84)  (10000, 84)};

\addplot[color=colk, mark=\markk] coordinates { (0.0,1) (0.03,2) (0.04,3) (0.05,4) (0.06,5) (0.07,6) (0.1,7) (0.13,8) (0.13,9) (0.15,10) (0.18,11) (0.23,12) (0.25,13) (0.39,14) (0.41,15) (0.42,16) (0.42,17) (0.6,18) (0.69,19) (0.73,20) (1.27,21) (1.36,22) (1.49,23) (2.13,24) (3.97,25) (4.27,26) (5.95,27) (6.38,28) (6.84,29) (7.49,30) (12.37,31) (14.94,32) (15.49,33) (15.81,34) (17.5,35) (33.7,36) (35.48,37) (39.78,38) (45.44,39) (54.45,40) (61.47,41) (109.42,42) (239.37,43) (264.92,44) (326.73,45) (351.02,46) (395.58,47) (424.29,48) (443.67,49) (465.61,50) (468.53,51) (582.08,52) (641.44,53) (778.61,54) (823.84,55) (887.18,56) (914.07,57) (1096.13,58) (1455.58,59) (1544.5,60) (2690.59,61) (3395.9,62) (3589.92,63) (4331.65,64) (4864.02,65)  (10000, 65)};

\legend{\solvere, \solverg, \solverh, \solverf, \solverc, \solverd, \solveri, \solverj, \solverb, \solvera, \solverk}

\addplot[densely dotted, color=black] coordinates {(5000, -1) (5000, 400)};

          \end{axis}
        \end{tikzpicture}
  
  \caption{\color{black} Solve time comparisons between base \yalsatprob{}, and 10 \yalsatddfw{} settings for \instancesA{} and \instancesB{} with seed 123, where restarts are enabled}
  \label{fig:comb+sc_res_large_seed123}
\end{figure}
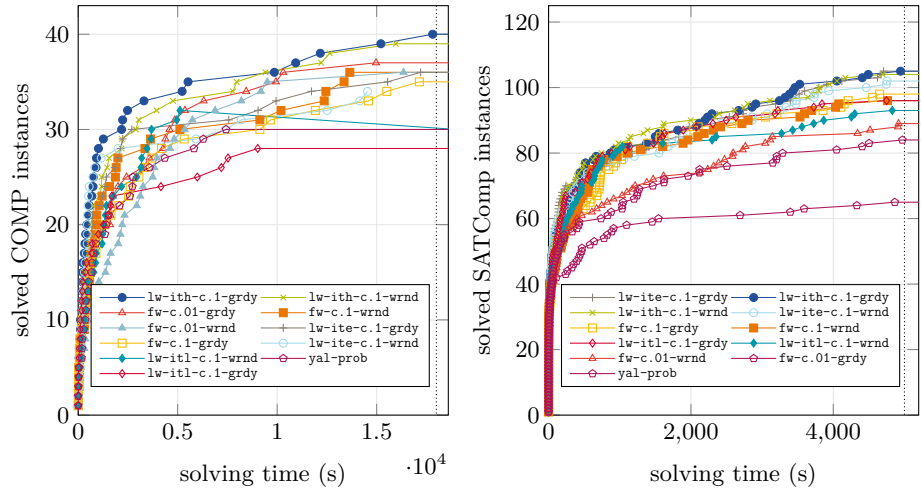
\end{subappendices}
\newpage
\bibliographystyle{splncs04}
\bibliography{citations}

\end{document}